\documentclass[]{interact}
\usepackage{booktabs}
\usepackage{tabularx}
\usepackage{booktabs}
\usepackage{tabularx}
\usepackage{array}
\usepackage{booktabs}
\usepackage{graphicx}
\usepackage{xcolor}
\usepackage{multirow}
\usepackage{pgfplots}
\usepackage{xcolor}
\usepackage{float}
\usepgfplotslibrary{groupplots}
\pgfplotsset{compat=1.18}
\usepackage{pgfplots}
\usepgfplotslibrary{groupplots}
\pgfplotsset{compat=1.18}
\usepackage{xcolor}

\definecolor{rawgray}{RGB}{140,140,140}
\definecolor{viterbiblue}{RGB}{35,80,120}

\usepackage{tikz}
\usepackage{pgfplots}
\usepgfplotslibrary{groupplots}
\pgfplotsset{compat=1.18}

\newcommand{\best}[1]{\textbf{#1}}

\usepackage{epstopdf}
\usepackage{subfigure}
\usepackage{float}
\usepackage{multirow}
\usepackage{makecell}
\usepackage{placeins}
\usepackage{graphicx}
\usepackage[sort]{natbib}
\bibpunct[, ]{(}{)}{;}{a}{}{,}

\theoremstyle{plain}

\theoremstyle{definition}

\theoremstyle{remark}

\begin{document}

\title{Delay-Guided Attention for Quality Prediction in Multistage Manufacturing}


\author{
\name{
Jiarui Zhang\textsuperscript{a,$\dagger$}\thanks{$\dagger$
Jiarui Zhang and Zhaoyang Guan contributed equally to this work.
},
Zhaoyang Guan\textsuperscript{a,$\dagger$},
Zhixuan Xia\textsuperscript{a},
and Naichen Shi\textsuperscript{b,$\ast$}\thanks{$\ast$
CONTACT Naichen Shi. Email: naichen.shi@northwestern.edu
}
}
\affil{
\textsuperscript{a}
Department of Engineering Sciences and Applied Mathematics,
Northwestern University, Evanston, IL, USA;\\
\textsuperscript{b}
Department of Industrial Engineering and Management Sciences,
Northwestern University, Evanston, IL, USA
}
}

\suppldatatrue
\maketitle
\suppldatafalse

\begin{abstract}
Quality prediction in multistage manufacturing is challenging when inter-stage lags vary over time, as measurements taken at the same timestamp may reflect different process states. Existing approaches often preserve production order but fuse timestamp-aligned features or model delayed dependencies implicitly, without identifying the upstream history relevant to each downstream observation. We propose Delay-Guided Attention for Multi-stage Forecasting (DAMF), which explicitly estimates inter-stage lag distributions and uses them to guide cross-stage information transfer for final quality prediction. For each adjacent stage pair, a lag predictor compares the current downstream representation with candidate lagged upstream representations. An occurrence--magnitude factorization separately estimates the probability of a positive lag and its conditional magnitude, yielding a sample- and stage-pair-specific distribution over discrete lag states without requiring plant-specified delays at inference. DelayAlignment incorporates this soft distribution as a temporal prior into differentiable cross-stage attention, while gated residual fusion regulates information propagation across production stages. For interpretation, segment-wise Viterbi decoding converts pointwise lag distributions into temporally coherent lag paths, whereas forecasting retains the soft distributions for alignment. On the real-world LiquidSugar dataset, DAMF improves final quality prediction over the compared baselines. Evaluations on two controlled-delay benchmarks show more accurate and stable lag recovery, and ablation studies support the contributions of lag prediction, delay-guided attention, and structured temporal decoding.
\end{abstract}

\begin{keywords}
Multistage manufacturing; Quality prediction; Lag recovery
\end{keywords}

\section{Introduction} 
Modern industrial production systems often comprise several physically or functionally connected stages in which raw materials are transformed into a series of intermediate products and the final product~\citep{Ceglarek2004SOVA,Du2015Markov,Filz2024Digitalization}. Although different stages perform different operations, their process states are not independent. Variations in equipment condition and operating settings at an upstream stage may propagate to downstream measurements and ultimately affect final product quality~\citep{yan2021deepmultistage,cho2024multistagenet,zhang2022pgat, Lawless1999Variation,
Agrawal1999Variation,Du2015Markov}. 

\begin{figure}[h]

    \centering
    \vspace{-0.2cm}
    \includegraphics[width=0.8\textwidth]{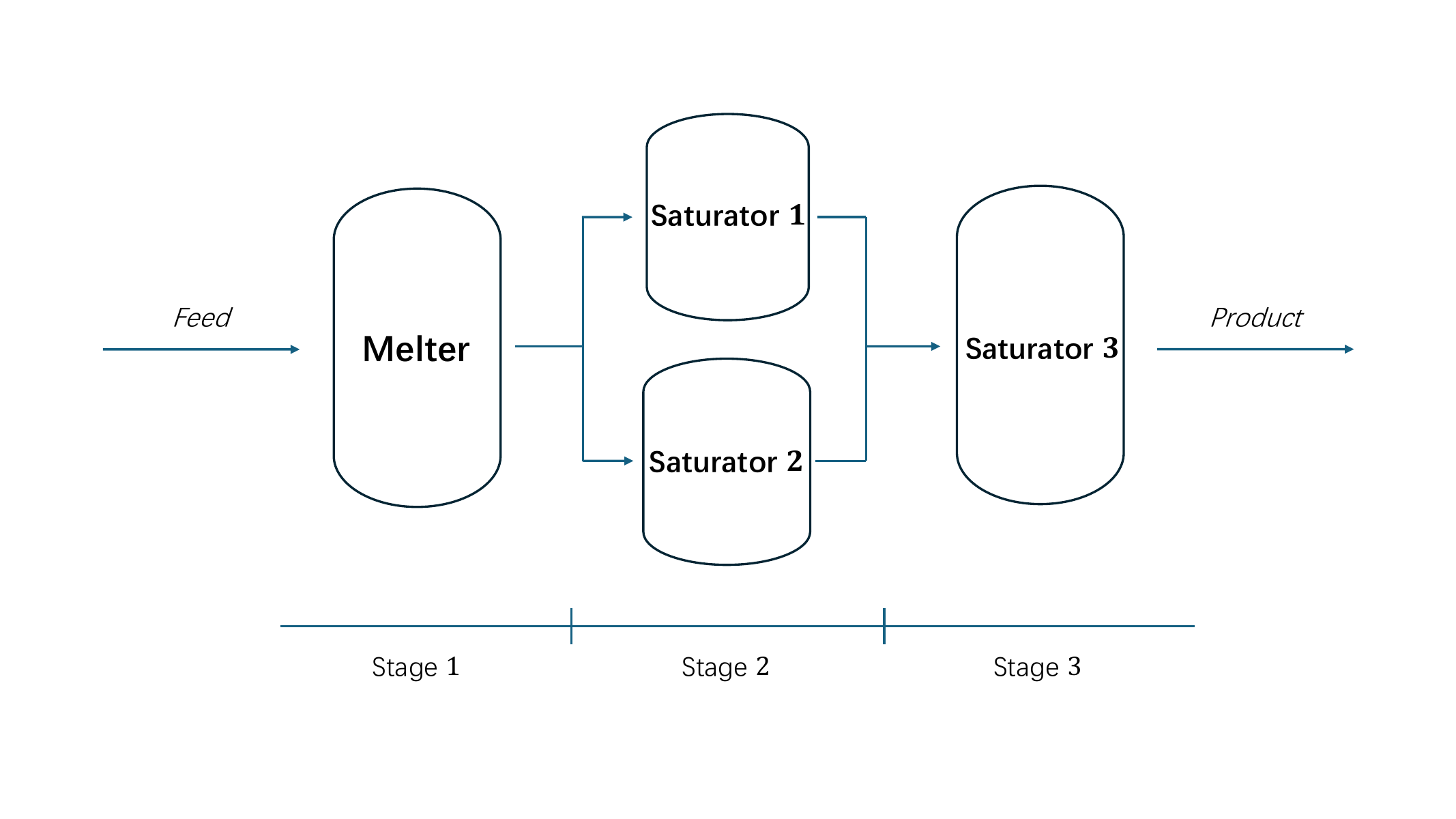}
    \vspace{-0.5cm}
    \caption{Schematic illustration of the sequential multi-stage Liquid Sugar production process.}
    \label{fig:liquidsugar-process}
\end{figure}

The liquid sugar production process is a representative example. As illustrated in Figure~\ref{fig:liquidsugar-process}, the feed first enters a melter in Stage~1. The resulting intermediate stream is processed by two parallel saturators in Stage~2, whose outputs are combined and transferred to Saturator~3 in Stage~3 before the final product is obtained~\citep{cho2024multistagenet}. Sensors associated with individual production stages continuously record stage-specific process variables, providing monitoring signals about the production process~\citep{Jiang2021SoftSensors,Qin2012ProcessMonitoring}.

The objective of this study is to predict the final product quality from the complete multi-stage sensor history with structural understandings that can support process diagnosis. We formulate this objective as two related tasks: forecasting the final quality of the manufacturing system output and describing the signal-propagation structure of the system. The two tasks are connected because final quality depends not only on the states recorded at individual stages, but also on which upstream historical patterns are associated with the current downstream condition. 

To understand the cross-stage signal propagation in multi-stage manufacturing, we study the signal lag. Although all measurements are often recorded under a common clock, equal timestamps do not necessarily refer to the same material condition or production event. Transportation, intermediate processing, and equipment response may cause a pattern recorded upstream to appear downstream after a temporal offset~\citep{Offermans2024SensorDelay,Cattaldo2024VariableDelay,Mejic2019DeadTime}. As a result, timestamp-based alignment is needed to associate a downstream observation with an upstream state~\citep{Offermans2020Synchronization,Yao2020VTR,Xiong2017TimeDelay}. Estimating this offset can also provide evidence about when and at which stage a quality-relevant deviation first becomes observable.

Accurate estimation of inter-stage lags is often challenging for several reasons. Different stage pairs involve different transportation and processing mechanisms, and their effective lags may vary with production rate, operating condition, equipment status, or material characteristics~\citep{Wang2021DynamicDelay,Zhao2021Cement,Reitz2013Residence}. A single global lag or static timestamp shift prescribed by domain knowledge is therefore unlikely to fully capture the dynamics of the multi-stage process~\citep{Yang2017DynamicDelay,GriesingScheiwe2020VariableDelay,Yao2020VTR}. Moreover, not every upstream variation remains equally informative after intermediate processing, as some patterns may be attenuated, whereas others become amplified. Another challenge is that lags are not directly observed and have been estimated from noisy observations. The statistical variations in monitoring signal complicates the estimation of lags across stages~\citep{Ni2010TimeDelay,Zheng1990TimeLag,Wang2024DNorm}. 


In literature, a few lag-detection and temporal causal-discovery methods study delayed relationships between time-series variables~\citep{runge2019detecting,pamfil2020dynotears}. However, these methods often model lag prediction as a task independent of quality prediction. When the two tasks are handled separately, prediction models do not benefit from sample-dependent lag distributions for different consecutive stage, which implicitly discards the interaction patterns across stages. 

Can we build a synergistic framework for multi-stage manufacturing signal analysis that transfers the information across the two tasks? Since lag prediction is often a local analysis between two adjacent manufacturing stages, while the quality prediction involves the comprehensive analysis of the entire system, combining them is not easy. To bridge the gap, we propose \emph{Delay-Guided Attention for Multi-stage Forecasting}(DAMF), which leverages the information from cross-stage lag estimation for final quality prediction. DAMF consists of two stages: lag identification and quality prediction. In the \textit{lag identification} stage, DAMF compares every current downstream representation with every representation from the upstream stage and estimates a data-dependent distribution of the lag. The resulting distribution serves as a prior in a differentiable DelayAlignment module, which calculates an attention map that prescribes a greater attention to upstream positions supported by the estimated lag. For lag recovery and interpretation, a Viterbi decoder converts the pointwise lag distributions into temporally coherent lag paths. Quality forecasting instead retains the soft distributions and uses them as priors in DelayAlignment. In the \textit{quality prediction} stage, the quality model inherits the structural information from lag prediction through cross-attention and outputs the final quality label. 

The main contributions of this work are summarized as follows:

\begin{enumerate}
    \item \textbf{Delay-aware multi-stage prediction framework.}
    We propose DAMF, a multi-stage quality prediction framework that jointly models inter-stage delays and predicts final product quality.

    \item \textbf{Inter-stage lag estimate.}
    We develop an inter-stage attention-based lag estimator that compares downstream representations with
    candidate upstream histories for each adjacent stage pair. The lag estimation module
    outputs both the occurrence of the lag and the magnitude of the lag, providing flexibility in industrial practice. We also propose a Viterbi filter to stabilize the lag estimates.

    \item \textbf{Lag-guided time series analysis.}
    We introduce a differentiable DelayAlignment module that incorporates the
    estimated lag distribution as a temporal prior to the attention map. Based on the attention map, we further propose a gated residual fusion
     module for final-quality prediction.

    \item \textbf{Experimental validation.}
    Experiments on the real LiquidSugar dataset and two controlled-delay benchmarks,
    together with systematic ablation studies, validate the effectiveness of DAMF
    for both final-quality prediction and inter-stage lag recovery.
\end{enumerate}

\section{Related Work}
\noindent\textbf{Multi-stage quality-prediction models.}
Data-driven soft sensors estimate difficult-to-measure quality variables from readily available process measurements~\citep{kadlec2009data}, but conventional soft-sensing models do not usually encode the ordered structure of a multi-stage manufacturing system. Several stage-aware models have been developed to address this limitation. The deep multi-stage multi-task learning model proposed by~\cite{yan2021deepmultistage} follows the sequential production structure and jointly predicts multiple stage-wise quality variables through shared representations.~\cite{wang2023production} use a multi-scale convolutional network with control gates to represent cumulative variation across stages and a multi-gate mixture-of-experts model to learn multiple quality tasks. PGAT~\citep{zhang2022pgat} represents machines and production paths as a directed graph and applies path-enhanced bidirectional graph attention to aggregate quality-relevant information. For continuous processes without intermediate quality measurements, MMPNet~\citep{cho2024mmpnet} sequentially injects stage-specific inputs into a feedforward network, while MultiStageNet~\citep{cho2024multistagenet} combines temporal encoders, interstage blocks, and stacked multi-stage modules to learn sequential dependencies without intermediate output or lead-time labels. These models capture stage-wise dependencies through multi-task learning, gated feature transfer, graph message passing, or sequential interstage modules. However, their cross-stage temporal relationships remain implicit: they do not expose a sample-dependent lag distribution for each adjacent stage pair and use it to align upstream and downstream observations.
\\
\\
\noindent\textbf{Time-series forecasting models.}
Deep time-series models provide alternative architectures for extracting temporal patterns from industrial sensor data. PatchTST~\citep{nie2023patchtst} divides each variable into temporal patches and applies channel-independent Transformer encoding, whereas iTransformer~\citep{liu2024itransformer} treats individual variables as tokens and learns their dependencies through attention. TimeMixer~\citep{wang2024timemixer} represents temporal variation at multiple resolutions and mixes information across scales. State-space models such as MambaTS~\citep{cai2024mambats} and TimeMachine~\citep{ahamed2024timemachine} provide efficient mechanisms for capturing long-range dependencies in multivariate sequences. These models are effective general-purpose forecasting architectures, but they treat the input primarily as a univariate or multivariate sequence. When applied directly to multi-stage manufacturing data, the known production order and adjacent-stage relationships are not explicitly used to determine which upstream historical position should be associated with a downstream observation. Temporal offsets between stages are therefore absorbed into the learned representations rather than represented as explicit, interpretable variables.
\\
\\
\noindent\textbf{Lag identification and structured temporal modeling.}
Methods for lagged-dependency analysis explicitly examine delayed relationships between time-series variables. Granger causality analysis~\citep{granger1969investigating} evaluates whether past observations of one variable improve the prediction of another. PCMCI~\citep{runge2019detecting} uses conditional-independence testing to identify lagged dependencies, while DYNOTEARS~\citep{pamfil2020dynotears} learns time-lagged directed structures through continuous optimization. These methods identify candidate lagged relationships under their respective assumptions, but their outputs are generally variable-level dependencies or temporal graphs estimated over a dataset or analysis window. They are not designed to produce a sample-dependent lag distribution for each manufacturing-stage pair and use that distribution directly in final-quality forecasting. Once point-wise lag scores are available, structured sequence methods can be used to improve temporal consistency. The Viterbi algorithm recovers the most probable state path from emission and transition scores~\citep{viterbi1967error,rabiner1989tutorial}, while explicit-duration models reduce unrealistically short assignments by modeling state persistence~\citep{yu2010hidden}. These decoders smooth an existing score sequence but do not estimate lag probabilities from cross-stage sensor features or connect them to the forecasting mechanism. DAMF links these previously separate components by estimating sample- and stage-pair-specific lag distributions, using them to guide cross-stage alignment for quality forecasting, and decoding them into coherent lag paths for interpretation.

\begin{figure}[!htbp]
\centering
\includegraphics[width=\textwidth]{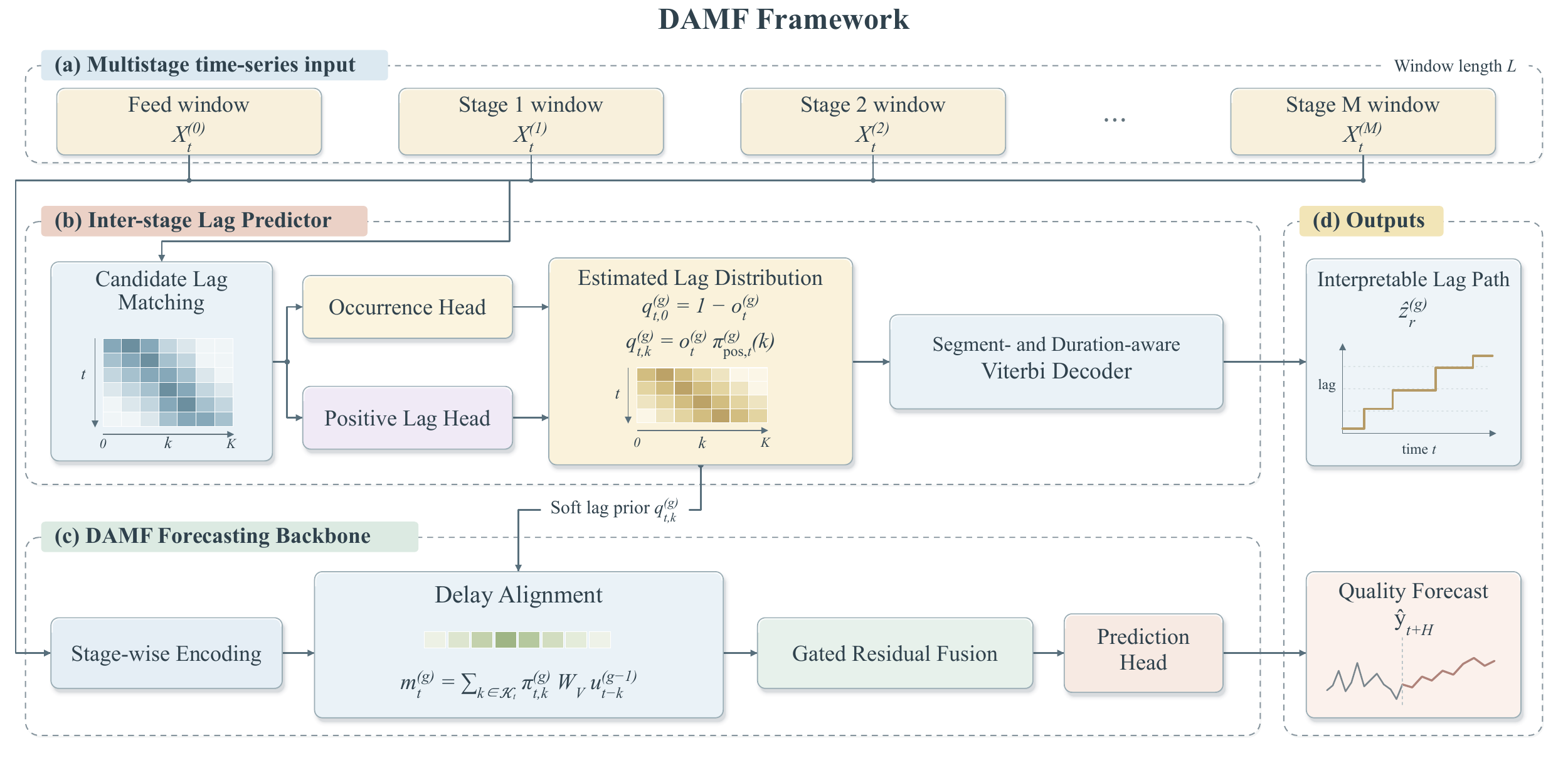}
\caption{Overall architecture of DAMF. Edge-wise lag distributions guide
DelayAlignment and gated residual fusion across successive processing
stages.}
\label{fig:damf-framework}
\end{figure}

\section{Methodology}
In this section, we will first introduce the modeling of inter-stage lag in the monitoring signal, and then present our methods to leverage the identified lag into quality prediction.
\subsection{Lag modeling and DAMF}

Consider a multi-stage manufacturing process with an ordered set of stages
$\mathcal{G}=\{0,1,\ldots,M\}$, where stage $0$ denotes the system input and
stages $1,\ldots,M$ denote successive processing stages. At time $t$, the
observation from stage $g$ is denoted by
$\mathbf{x}_t^{(g)} \in \mathbb{R}^{d_g}$, where $d_g$ is the number of
variables measured at that stage. Given a window length $L$, the historical
input from stage $g$ is
\[
\mathbf{X}_t^{(g)}
=
\left[
\mathbf{x}_{t-L+1}^{(g)},
\mathbf{x}_{t-L+2}^{(g)},
\ldots,
\mathbf{x}_t^{(g)}
\right]^{\top}
\in \mathbb{R}^{L \times d_g}.
\]

\begin{figure}[!htbp]
\centering
\includegraphics[width=0.88\textwidth]
{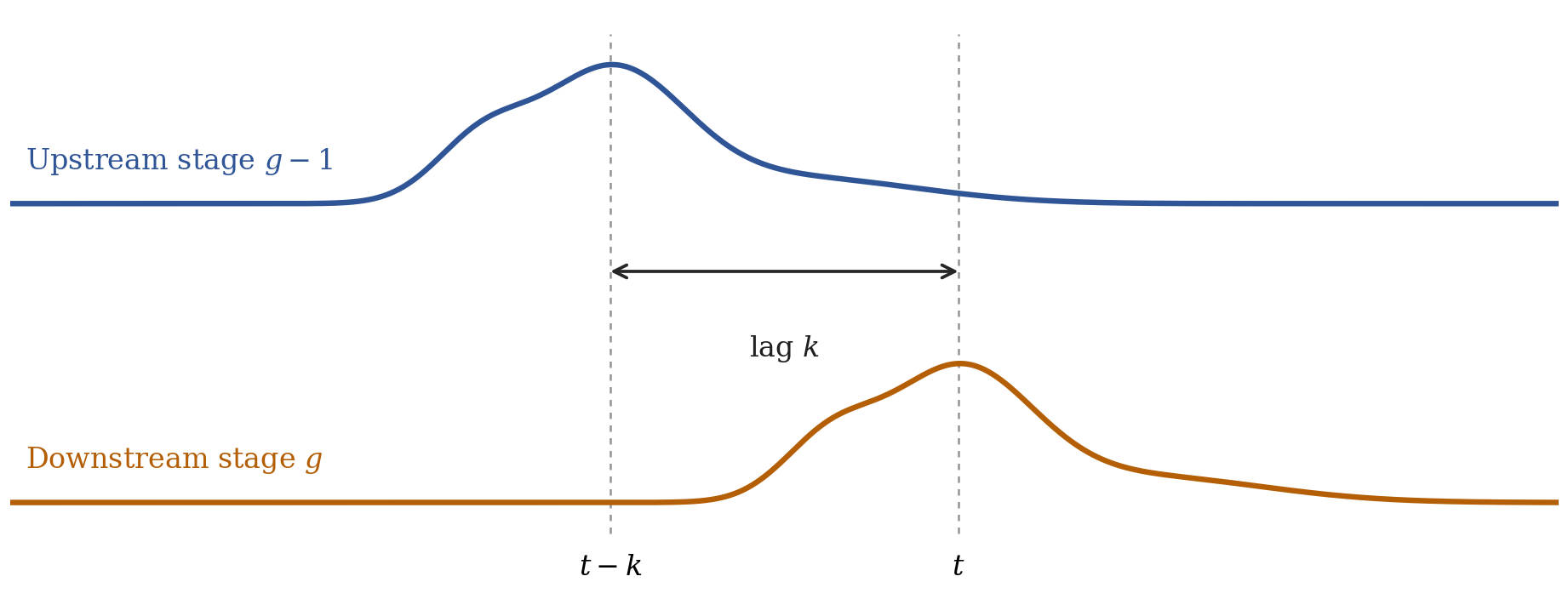}
\caption{Illustration of an inter-stage lag. An upstream process feature
observed at \(t-k\) corresponds to a downstream feature observed at \(t\),
giving a lag of \(k\) sampling steps.}
\label{fig:lag-illustration}
\end{figure}

Let \(y_t\) denote the final-quality response at time \(t\).
Using the multi-stage history
\(\{\mathbf{X}_t^{(g)}\}_{g=0}^{M}\), we aim to predict
\(y_{t+H}\), where \(H\) is the forecasting horizon measured
in sampling steps. Although the stage-wise signals share a common time index,
corresponding process states may not capture the same physical process at the same time. We represent this temporal
offset by a discrete lag state between two adjacent stage $(g-1,g)$ for $g=1,\ldots,M$ as:
\[
z_t^{(g)} \in \mathcal{K},
\qquad
\mathcal{K}=\{0,1,\ldots,K\},
\]
where $K$ is the maximum candidate lag. The state $z_t^{(g)}=k$ indicates
that the downstream state at time $t$ is aligned with an upstream state
observed at time $t-k$. Here, the lag represents a signal-level alignment
offset. Figure~\ref{fig:lag-illustration} illustrates a lag of $k$ sampling steps between two
adjacent stages.

Based on this formulation, we propose Delay-Guided Attention for
Multi-stage Forecasting (DAMF), which consists of inter-stage lag
estimation and lag-guided forecasting. For each pair of adjacent stages $(g-1,g)$, the
lag predictor estimates the posterior probability distribution of the lag $z_t^{(g)}$ after observing the signal from the two stages,
\[
q_{t,k}^{(g)}
=
\Pr\left(
z_t^{(g)}=k
\mid
\mathbf{X}_t^{(g-1)},\mathbf{X}_t^{(g)}
\right),
\qquad k\in\mathcal{K}.
\]
DelayAlignment combines this distribution with content-based matching to
construct an aligned upstream message, which is then integrated with the
current-stage representation through gated residual fusion. Repeating
this operation across successive stages constructs the representation needed for the final quality forecast \(\widehat{y}_{t+H}\). Figure~\ref{fig:damf-framework}
summarizes the overall architecture, and the following subsections describe
the two components in detail.

\subsection{Inter-stage lag estimation}
\label{sec:lag-estimation}

As discussed, for each adjacent-stage edge \((g-1,g)\), the lag predictor maps the
upstream and downstream sensor observations to a pointwise distribution
over the candidate lags. This subsection introduces the lag-matching
mechanism and the training strategy, then presents an inference strategy used to recover a temporally smooth lag path. The
pointwise distribution of the lag is subsequently
used for lag-guided forecasting.

\subsubsection{Lag matching and estimation}
\label{sec:lag-matching}

For an adjacent-stage edge \((g-1,g)\), we refer to
\(\mathbf{X}_t^{(g-1)}\) as the source window and
\(\mathbf{X}_t^{(g)}\) as the target window. Both contain the most
recent \(L\) observations up to time \(t\).
We encode both windows using gated recurrent unit (GRU)
networks~\citep{cho2014learning},
whose gates control how information from previous observations
is retained and updated. Intuitively, a GRU maintains a hidden state
that serves as a running summary of the sequential observation.
At each time step, a reset gate determines how much of the previous
summary is used to inform the current hidden state. Then, an update gate interpolates between the previous hidden state
and the current one. This gating mechanism
allows the network to identify persistent sequential patterns over the history. We leverage the flexibility
and denoising capability of GRU to extract important features from raw sensing measurements.

The output of GRU from source window is a
sequence of hidden states
\(\{\mathbf{h}_{t,j}^{(g-1)}\}_{j=1}^{L}\) representing the latent features extracted, where position \(j\)
corresponds to time \(t-L+j\). On the target side, the GRU
similarly summarizes the latent information from the corresponding time window into a context vector \(\mathbf{c}_t^{(g)}\).
For candidate lag \(k\), \(\mathbf{h}_{t,L-k}^{(g-1)}\) therefore
incorporated the source semantic information corresponding to time \(t-k\).
We assume \(K\leq L-1\).

To model the semantic correspondence between the target context and the source
representation under candidate lag \(k\), we propose a similarity metric denoted by,
\begin{equation}
s_{t,k}^{(g)}
=
\frac{
\left(W_m\mathbf{h}_{t,L-k}^{(g-1)}\right)^{\top}
\mathbf{c}_t^{(g)}
}{
\sqrt{d_h}
}
+b_k,
\qquad k\in\mathcal{K},
\label{eq:lag-matching-score}
\end{equation}
where \(W_m\) is a projection matrix that maps the source representation
to the space of the target representation, \(d_h\) is the dimension of this
space, and \(b_k\) is a lag-specific learnable bias. Ideally, a larger
\(s_{t,k}^{(g)}\) indicates stronger correspondence under lag \(k\).

After calculating the matching scores $s^{(g)}_{t,k}$ for all candidate lags $k$, we pass them into two prediction
heads that estimate lag occurrence and magnitude.
The occurrence head will output
\(o_t^{(g)}\), which represents the probability that there is a non-zero lag \(z_t^{(g)}>0\).
For magnitude estimation, the positive-lag head produces \(K\) outputs,
one for each lag \(k=1,\ldots,K\), to predict probability that the lag is $k$. Combining both outputs yields a
categorical distribution over all candidate lags:
\begin{equation}
q_{t,0}^{(g)}
=
1-o_t^{(g)},
\qquad
q_{t,k}^{(g)}
=
o_t^{(g)}\pi_{\mathrm{pos},t}^{(g)}(k),
\quad k=1,\ldots,K.
\label{eq:combined-lag-distribution}
\end{equation}
One can verify that the vector
\(\mathbf{q}_t^{(g)}=[q_{t,0}^{(g)},\ldots,q_{t,K}^{(g)}]\)
defines a normalized distribution over \(\mathcal{K}\), since the
positive-lag probabilities sum to one.

\subsubsection{Lag-predictor training}
\label{sec:lag-predictor-training}

After defining the parametric model to predict the occurrence of the lag and its distribution, we will introduce the training objectives to fit these models. 

Remember that primary task of DAMF is to recover the signal delay dynamics and predict the quality label simultaneously. We therefore design two main empirical loss functions following the hierarchical factorization
in Eq.~\eqref{eq:combined-lag-distribution}. The \textit{occurrence loss}
\(\mathcal{L}_{\mathrm{occ}}\) encourages the lag occurrence prediction \(o_t\) to match the ground truth binary label. A focal-style objective is used to reduce
the influence of the class imbalance in the training dataset, as most observations do not admit lags. Within the
positive-lag subset, the \textbf{magnitude loss}
\(\mathcal{L}_{\mathrm{pos}}\) is a cross-entropy loss that encourages
\(\pi_{\mathrm{pos},t}\) to identify the correct ground-truth lag among
\(\{1,\ldots,K\}\).

To further strengthen the robustness of the prediction, we introduce three auxiliary objectives. The
relaxed Top-\(N\) loss \(\mathcal{L}_{\mathrm{TopN}}\) operates on the
matching scores \(s_{t,k}\) and encourages the ground-truth lag
\(\ell_t\) to remain among the leading \(N\) candidates without requiring
it to be ranked first. The large-lag loss
\(\mathcal{L}_{\mathrm{large}}\) emphasizes the positive-occurrence
supervision for samples satisfying
\(\ell_t\geq K_{\mathrm{large}}\), reducing the tendency of large lags to
collapse to the zero-lag state. Finally, for samples whose ground-truth
lag is retained in the Top-\(N\) candidate set \(\mathcal{C}_t\), the
selector loss \(\mathcal{L}_{\mathrm{sel}}\) uses the selector
probabilities \(a_t(j)\) to refine the ranking among the retained
candidates. The selector provides auxiliary supervision only and is not
used to construct \(\mathbf{q}_t^{(g)}\) at inference.

The complete lag-predictor objective, denoted by
\(\mathcal{L}_{\mathrm{lag}}\), combines
\(\mathcal{L}_{\mathrm{occ}}\),
\(\mathcal{L}_{\mathrm{pos}}\), and
\(\mathcal{L}_{\mathrm{TopN}}\), together with the weighted auxiliary
terms \(\mathcal{L}_{\mathrm{large}}\) and
\(\mathcal{L}_{\mathrm{sel}}\). Their complete mathematical definitions
are provided in Appendix~\ref{app:lag-losses}. 

\subsubsection{Denoised inference of the lag}
\label{sec:lag-path-decoding}
Even with well-trained lag predictors, we notice that the direct lag predictions tend to fluctuate over time. We attribute such fluctuations to the stochasticity of the sensing data and training dynamics.
To encourage temporal consistency among lag predictions, we propose to use a Viterbi algorithm to smooth the fluctuations.

\begin{figure}[!htbp]
\centering
\includegraphics[width=0.62\textwidth]
{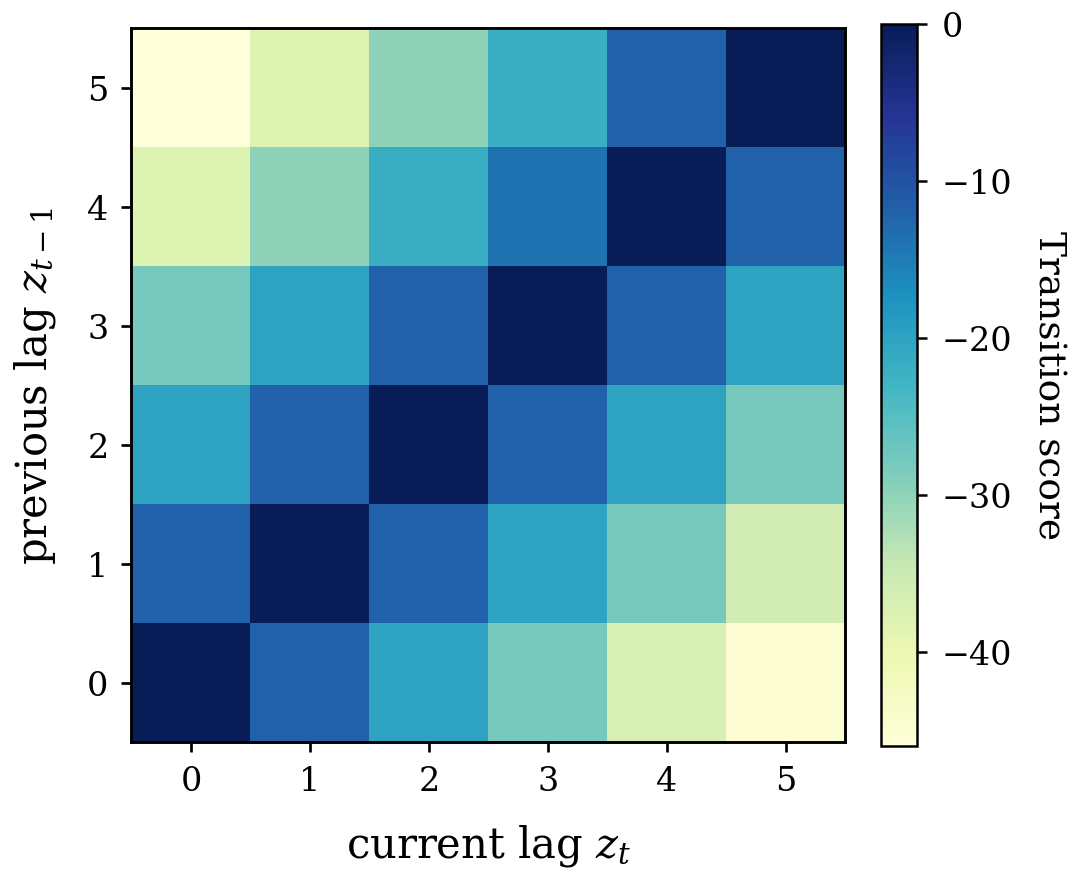}
\caption{Transition-score structure for lag-path decoding under the
experimental parameter settings. Darker cells indicate higher scores and smaller penalties.
}
\label{fig:viterbi-transition}
\end{figure}

More specifically, we use \(\mathcal{T}_r=(t_1,\ldots,t_{n_r})\) denote the \(n_r\) time
indices within a time window, ordered chronologically.
For the manufacturing stage \((g-1,g)\), a candidate path
\(\mathbf{z}_r^{(g)}\in\mathcal{K}^{n_r}\) assigns a lag state
\(z_{t_i}^{(g)}\) to each time \(t_i\).
Here, \(\mathcal{K}^{n_r}\) denotes the set of all lag sequences
of length \(n_r\) with entries in \(\mathcal{K}\).
We obtain the decoded path by maximizing the regularized score:
\begin{equation}
\hat{\mathbf{z}}_r^{(g)}
=
\arg\max_{\mathbf{z}_r^{(g)}\in\mathcal{K}^{n_r}}
\left\{
\sum_{i=1}^{n_r}
\log\left(
q_{t_i,z_{t_i}^{(g)}}^{(g)}+\epsilon
\right)
+
\sum_{i=2}^{n_r}
A\left(
z_{t_{i-1}}^{(g)},z_{t_i}^{(g)}
\right)
\right\},
\label{eq:lag-path-decoding}
\end{equation}
where \(\epsilon>0\) is a small constant added for numerical stability,
and \(A(a,b)\) denotes the transition score from lag state \(a\)
to lag state \(b\).
In this objective, the first sum measures agreement between the selected lag and the
pointwise predicted lag scores, while the second penalizes abrupt changes between successive lag states.
For a transition from state \(a\) to state \(b\), with
\(a,b\in\mathcal{K}\), we define
\begin{equation}
\begin{aligned}
A(a,b)
={}&-\eta_{\mathrm{sw}}\mathbb{I}(a\neq b)
-\eta_{\mathrm{jump}}|a-b| \\
&-\eta_{\mathrm{drop}}
\mathbb{I}(a\geq\kappa \,\wedge\, 0<b<\kappa) \\
&-\eta_{\mathrm{off}}
\mathbb{I}(a\geq\kappa \,\wedge\, b=0)
-\eta_{\mathrm{on}}
\mathbb{I}(a=0 \,\wedge\, b\geq\kappa).
\end{aligned}
\end{equation}
Here, \(\kappa\in\{1,\ldots,K\}\) is the threshold defining
large-lag states, and \(\mathbb{I}(\cdot)\) is the indicator function.
All penalty weights are nonnegative.
The coefficients \(\eta_{\mathrm{sw}}\) and
\(\eta_{\mathrm{jump}}\) penalize state switching and the magnitude
of a lag change, respectively. Additional penalties
\(\eta_{\mathrm{drop}}\), \(\eta_{\mathrm{off}}\), and
\(\eta_{\mathrm{on}}\) apply to transitions from a large lag to
a smaller positive lag, from a large lag to zero, and from zero
to a large lag, respectively.
Because the first two terms depend only on \(|a-b|\), they define
a symmetric Toeplitz base transition matrix. The remaining terms
introduce transition-specific corrections.
Figure~\ref{fig:viterbi-transition} illustrates the resulting transition
structure.

We use the decoded path only for lag recovery and interpretation.
For quality forecasting, DAMF instead uses the pointwise distribution
\(\mathbf{q}_t^{(g)}\) as a soft prior in DelayAlignment.

\subsection{Lag-guided forecasting}

When the lag distribution predictors are optimized, we use these predictions as alignment priors to predict the final quality label.
For this quality prediction task, we similarly employ GRU to encode the temporal observations into semantic embeddings. For stage $g$, we use \(\mathbf{e}_t^{(g)}\) to denote the semantic representation at
time \(t\), and let \(\mathbf{u}_t^{(g)}\)
to denote the representation of the upstream information.
At the system input, we set
\(\mathbf{u}_t^{(0)}=\mathbf{e}_t^{(0)}\).

For two adjacent stages \((g-1,g)\), the DelayAlignment module compares
\(\mathbf{e}_t^{(g)}\) with upstream representations at candidate
lagged positions.  For each valid lag \(k\), we construct the cross-attention score as
\begin{equation}
\widetilde{\alpha}_{t,k}^{(g)}
=
\frac{
\left\langle
W_Q\mathbf{e}_t^{(g)},
W_K\left(\mathbf{u}_{t-k}^{(g-1)}+\mathbf{r}_k\right)
\right\rangle
}{
\sqrt{d_a}
}
+\beta_k
+\lambda\log\left(q_{t,k}^{(g)}+\epsilon\right).
\label{eq:lag-guided-alignment}
\end{equation}
Here, \(W_Q\) and \(W_K\) project the downstream and upstream
representations into query and key vectors of dimension \(d_a\),
and \(\langle\cdot,\cdot\rangle\) denotes their inner product.
The vector \(\mathbf{r}_k\) is an embedding of lag \(k\), and
\(\beta_k\) is a lag-specific learnable bias in DelayAlignment.
The coefficient \(\lambda\) controls the contribution of the lag
prior, while \(\epsilon>0\) ensures numerical stability.
Setting \(\lambda=0\) removes the lag prior.

Applying softmax to
\(\widetilde{\alpha}_{t,k}^{(g)}/\tau\) over
\(k\) gives the alignment weights
\(\pi_{t,k}^{(g)}\), where \(\tau>0\) is the temperature parameter.
These weights depend on both the encoded representations and the
estimated lag distribution. We use them to aggregate the upstream
information into a message:
\begin{equation}
\mathbf{m}_t^{(g)}
=
\sum_{k\in\mathcal{K}_t}
\pi_{t,k}^{(g)}W_V\mathbf{u}_{t-k}^{(g-1)},
\label{eq:upstream-message}
\end{equation}
where \(W_V\) is the value projection matrix.
This weighted aggregation allows multiple candidate lags to
contribute, rather than selecting a single upstream position.

\begin{figure}[h!]
    \centering
    \includegraphics[width=\linewidth]{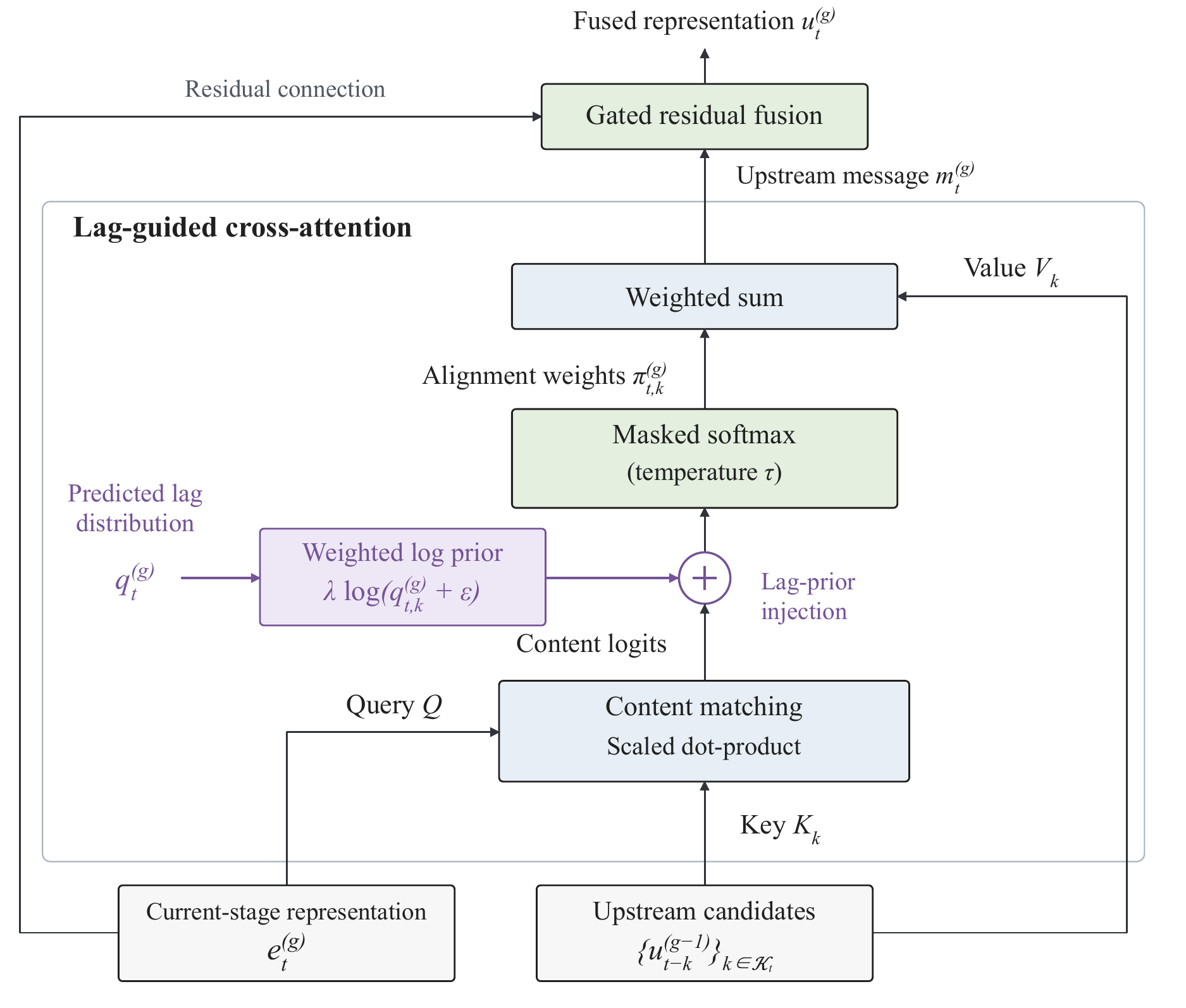}
     \caption{Lag-guided alignment and fusion for an adjacent-stage
    pair. DelayAlignment combines content-based matching with the
    estimated lag distribution to compute attention weights over
    upstream candidates. The resulting upstream message is combined
    with the current-stage representation through gated residual
    fusion.}
    \label{fig:lag-guided-forecasting}
\end{figure}

To combine the aligned message with the current-stage representation,
we introduce two nonlinear information fusion mechanisms:
\begin{equation}
\begin{aligned}
\gamma_t^{(g)}
&=
\sigma\left(
W_\gamma^{(g)}
[\mathbf{e}_t^{(g)};\mathbf{m}_t^{(g)}]
+\mathbf{b}_\gamma^{(g)}
\right), \\
\mathbf{u}_t^{(g)}
&=
\mathbf{e}_t^{(g)}
+
\gamma_t^{(g)}\odot
\psi^{(g)}\left(\mathbf{m}_t^{(g)}\right),
\qquad g=1,\ldots,M.
\end{aligned}
\label{eq:gated-fusion}
\end{equation}
Here, \(\sigma\) is the nonlinear activation function such as a sigmoid function,
\([\cdot;\cdot]\) denotes vector concatenation, and
\(W_\gamma^{(g)}\) and \(\mathbf{b}_\gamma^{(g)}\) are
learnable gate parameters.
The mapping \(\psi^{(g)}\) transforms the message to the dimension
of \(\mathbf{e}_t^{(g)}\), and \(\odot\) denotes elementwise
multiplication. This update retains the current-stage representation
while adding a gated contribution from upstream. Figure~\ref{fig:lag-guided-forecasting} illustrates the alignment
and fusion operations for an adjacent-stage pair.

Repeating the update across successive stages produces
\(\mathbf{u}_t^{(M)}\).
A prediction head \(\phi_y\) then gives the final-quality forecast
\(\hat{y}_{t+H}=\phi_y(\mathbf{u}_t^{(M)})\).

Training proceeds in two phases with separate objectives.
We first train the lag predictor with
\(\mathcal{L}_{\mathrm{lag}}\), as described in Section~3.2.2.
After completing lag-predictor training, we use its estimated
distributions as alignment priors and train the forecasting model.
The objective for this second phase is the mean absolute error (MAE) between observed and predicted quality:
\begin{equation}
\mathcal{L}_y
=
\frac{1}{N_y}
\sum_{i=1}^{N_y}
\left|y_{i+H}-\hat{y}_{i+H}\right|,
\label{eq:forecasting-loss}
\end{equation}
where \(N_y\) is the number of forecasting training samples.

During forecasting training, the lag predictor is held fixed and
provides the pointwise distributions $q_t^{(g)}$ as soft alignment
priors. The forecasting loss $L_y$ updates only the stage encoders
in the forecasting backbone, DelayAlignment, the gated fusion
modules, and the prediction head.

\section{Experiments}
This section evaluates the proposed DAMF framework from three perspectives. First, we evaluate whether the proposed lag-recovery pipeline can recover meaningful inter-stage lag patterns on controlled-delay benchmarks. Second, we examine whether delay-integrated modeling improves final quality prediction on real multi-stage manufacturing data. Third, we conduct ablation studies to analyze the effects of lag-predictor architecture, loss components, structured temporal decoding, and delay priors.

\subsection{Datasets}
\label{sec:datasets}

We evaluate DAMF using one real-world manufacturing dataset and two controlled-delay benchmarks. The original LiquidSugar dataset is used for final quality prediction, while a delay-injected version of LiquidSugar and a CausalAI-based synthetic benchmark are used for inter-stage lag recovery. LiquidSugar~\citep{cho2024multistagenet} contains 22,854 observations with a nominal sampling interval of 15 minutes. Each observation consists of one timestamp, 24 process variables, and one final quality target, \texttt{yield\_flow}. The process variables include seven feed variables, two Stage-1 variables, twelve Stage-2 variables, and three Stage-3 variables. Observations separated by more than 120 minutes are treated as different continuous segments. Within each segment, measurements are regularized onto a 15-minute grid, and short missing intervals are filled by linear interpolation. The resulting complete segments are divided chronologically into training and test sets using an approximate 80/20 ratio, with identical preprocessing and data splits used for all compared methods.

For controlled lag-recovery evaluation, we further construct a delay-injected version of LiquidSugar. Within each split, 50\% of the eligible segments are randomly selected for lag injection, while the remaining segments retain the zero-lag condition. The candidate lag states are $\{0,1,2,3,4,5\}$ sampling steps, corresponding to physical delays ranging from 0 to 75 minutes in 15-minute increments. The injected lag sequence is retained as the ground-truth trajectory at each time step and is used to evaluate the accuracy of inter-stage lag recovery.

We additionally construct a CausalAI-based synthetic benchmark to evaluate lag recovery under fully controlled temporal dependencies. The benchmark consists of a bivariate time series with one source variable and one target variable, where the target is generated from the source under a prescribed time-varying lag. For this benchmark, the candidate lag states are $\{0,1,2,3,4,5,6\}$, with an 80/20 train/test split and lag changes introduced into 50\% of the eligible regions. Both block-shaped and smooth bump-shaped lag trajectories are included to represent persistent and transient delay variations, respectively. Since the ground-truth lag trajectory $g(t)$ is explicitly specified during data generation, this benchmark enables direct evaluation of both lag-recovery accuracy and the ability to track time-varying delay dynamics.

\subsection{Experimental Settings}
\label{sec:experimental-settings}
The experiments address two tasks: final quality prediction and inter-stage lag recovery. All methods follow the data splits and preprocessing procedures described in Section~\ref{sec:datasets}. For final quality prediction, DAMF is compared with three groups of baselines.
The machine-learning baselines include Linear Regression,
XGBoost~\citep{chen2016xgboost}, Random Forest~\citep{breiman2001random}, and LightGBM~\citep{ke2017lightgbm}. The deep sequential baselines include LSTM~\citep{hochreiter1997long}, GRU~\citep{cho2014learning}, and Transformer~\citep{vaswani2017attention}. The  multistage baselines include TimeXer~\citep{wang2024timexer}, PEGNN-SS~\citep{niresi2024pegnn}, and MultistageNet~\citep{cho2024multistagenet}.

These three baselines are selected for their complementary modeling capabilities. TimeXer captures temporal dependencies and
exogenous-variable interactions, providing a reference for attention-based prediction. PEGNN-SS incorporates physical knowledge into sensor graphs, representing physics-enhanced modeling of process-variable interactions. MultistageNet learns sequential stage dependencies and lead-time relationships without intermediate output or lead-time labels, making it directly relevant to our manufacturing setting. Together, they provide complementary comparisons for evaluating DAMF's explicit modeling of inter-stage delays. Lag recovery is evaluated separately on the controlled-delay versions of LiquidSugar and the CausalAI-based synthetic benchmark. Both benchmarks contain block-shaped and smooth bump-shaped lag patterns, representing persistent and transient delay changes, respectively.

For lag-recovery evaluation, DAMF is compared with MNet and WCC. MNet uses the hidden lag predictions produced by MultistageNet~\citep{cho2024multistagenet} as its lag estimates. WCC applies window-based cross-correlation, a classical time-delay estimation approach~\citep{knapp1976generalized,carter1987coherence}, followed by a conventional Viterbi decoder.

The input window length is set to $L=12$ throughout the experiments. For lag recovery, the maximum candidate lag is set to $K=5$ for LiquidSugar and $K=6$ for CausalAI. The lag predictors use two-layer GRU encoders with a hidden dimension of 64 and are trained for 60 epochs using Adam with a learning rate of $5\times10^{-4}$. The batch sizes are 256 and 64 for LiquidSugar and CausalAI, respectively.

For final quality prediction on LiquidSugar, we use a forecasting horizon of $H=1$. The DAMF forecasting backbone also employs two-layer GRU encoders with a hidden dimension of 64. It is trained for 50 epochs using Adam with a learning rate of $10^{-3}$ and a batch size of 256, while the pretrained lag predictor remains fixed.

\subsubsection{Evaluation metrics}
\label{sec:evaluation_metrics}
Final quality prediction is evaluated using RMSE, MAE, and $R^2$, where lower RMSE and MAE and higher $R^2$ indicate better performance. For lag recovery, let $\mathcal{T}$ denote the set of evaluated test indices, and let $d_t$ and $\hat d_t$ denote the true and predicted lags, respectively, measured in sampling steps. We define $\mathcal{I}=\{t\in\mathcal{T}:d_t>0\}$,$\mathcal{O}=\{t\in\mathcal{T}:d_t=0\}$, and
$\mathcal{L}=\{t\in\mathcal{T}:d_t\geq4\}$ as the sets of true positive-lag, zero-lag, and large-lag test points, respectively.

\noindent\textbf{Lag occurrence.}
For the controlled-delay comparison, let $\mathrm{TP}_{d}$ and
$\mathrm{FP}_{d}$ count positive-lag detections in $\mathcal{I}$ and
$\mathcal{O}$, respectively. For DAMF, a positive lag is detected when
$\hat d_t>0$. Precision and the false alarm rate (FAR) are defined as
\begin{equation}
\mathrm{Precision}
=\frac{\mathrm{TP}_{d}}{\mathrm{TP}_{d}+\mathrm{FP}_{d}},
\qquad
\mathrm{FAR}
=\frac{\mathrm{FP}_{d}}{|\mathcal{O}|}.
\label{eq:eval_detection}
\end{equation}
These metrics assess whether a lag is detected, rather than whether its
magnitude is estimated exactly. For the lag-predictor ablations, occurrence F1 (Occ F1) uses the predicted
occurrence probability $o_t$. A ground-truth label is positive
when $d_t>0$, and a predicted label is positive when $o_t\geq0.5$.
Let $\mathrm{TP}_{o}$, $\mathrm{FP}_{o}$, and $\mathrm{FN}_{o}$ denote the
true-positive, false-positive, and false-negative counts obtained from
these labels over all test points. Then
\begin{equation}
\text{Occ F1}
=\frac{2\mathrm{TP}_{o}}
{2\mathrm{TP}_{o}+\mathrm{FP}_{o}+\mathrm{FN}_{o}}.
\label{eq:eval_occurrence_f1}
\end{equation}

\noindent\textbf{Lag magnitude and peak recovery.}
The mean absolute error over true positive-lag points is reported as Lag MAE in the ablations and In-bump MAE in the controlled-delay comparison.
Outside mean delay measures the average predicted lag at true zero-lag points. These metrics are defined as
\begin{equation}
\begin{aligned}
\text{Lag MAE}
&=\frac{1}{|\mathcal{I}|}
\sum_{t\in\mathcal{I}}|\hat d_t-d_t|,\\[3pt]
\text{Outside mean delay}
&=\frac{1}{|\mathcal{O}|}
\sum_{t\in\mathcal{O}}\hat d_t.
\end{aligned}
\label{eq:eval_lag_magnitude}
\end{equation}
For peak value error, let $\mathcal{I}_b$, $b=1,\ldots,B$, denote the maximal contiguous positive-lag intervals within individual data segments. We define
\begin{equation}
\text{Peak Value Error}
=\frac{1}{B}\sum_{b=1}^{B}
\left|
\max_{t\in\mathcal{I}_b}\hat d_t
-\max_{t\in\mathcal{I}_b}d_t
\right|.
\label{eq:eval_peak_error}
\end{equation}
This metric compares peak heights rather than peak timing; the predicted and true maxima need not occur at the same time point.

\noindent\textbf{Pointwise accuracy.}
Exact-match accuracy (Acc) and one-step-tolerance accuracy ($\pm1$ Acc) are computed over all test points, including zero-lag points:
\begin{equation}
\begin{aligned}
\mathrm{Acc}
&=\frac{|\{t\in\mathcal{T}:\hat d_t=d_t\}|}
{|\mathcal{T}|},\\[3pt]
\text{$\pm1$ Acc}
&=\frac{|\{t\in\mathcal{T}:|\hat d_t-d_t|\leq1\}|}
{|\mathcal{T}|}.
\end{aligned}
\label{eq:eval_lag_accuracy}
\end{equation}

\noindent\textbf{Large-lag recovery.}
The remaining three metrics are evaluated only over the true large-lag set $\mathcal{L}$:
\begin{equation}
\begin{aligned}
\text{Large-lag Recall}
&=\frac{|\{t\in\mathcal{L}:\hat d_t\geq4\}|}
{|\mathcal{L}|},\\[3pt]
\text{Collapse to 0}
&=\frac{|\{t\in\mathcal{L}:\hat d_t=0\}|}
{|\mathcal{L}|},\\[3pt]
\text{Collapse to small}
&=\frac{|\{t\in\mathcal{L}:1\leq\hat d_t\leq3\}|}
{|\mathcal{L}|}.
\end{aligned}
\label{eq:eval_large_lag}
\end{equation}
Large-lag Recall measures whether a true large lag is still predicted as large, without requiring an exact match. Collapse to 0 and Collapse to small measure the fractions of true large lags underestimated as zero and as positive lags of one to three sampling steps, respectively. Higher precision, Occ F1, Acc, $\pm1$ Acc, and Large-lag Recall indicate better performance; lower values are preferred for the other lag-recovery metrics.
\subsection{Main Results}
\begin{table}[!h]
\centering
\tbl{Final quality prediction performance on the LiquidSugar dataset. The best result for each metric is highlighted in bold.}
{
\footnotesize
\setlength{\tabcolsep}{6pt}
\renewcommand{\arraystretch}{1.12}
\begin{tabular}{@{}llccc@{}}
\toprule
\multirow{2}{*}{Category} & \multirow{2}{*}{Method} & \multicolumn{3}{c}{LiquidSugar} \\
\cmidrule(lr){3-5}
& & RMSE $\downarrow$ & MAE $\downarrow$ & $R^2 \uparrow$ \\
\midrule
\multirow{7}{*}{Machine Learning Methods}
& Linear Regression & 8.86 $\pm$ 0.00 & 6.23 $\pm$ 0.00 & 0.06 $\pm$ 0.00 \\
& XGBoost & 6.12 $\pm$ 0.05 & 4.52 $\pm$ 0.11 & 0.55 $\pm$ 0.01 \\
& LSTM & 10.24 $\pm$ 3.58 & 7.43 $\pm$ 1.92 & $-0.36 \pm 0.97$ \\
& Random Forest & 6.71 $\pm$ 0.05 & 5.06 $\pm$ 0.06 & 0.46 $\pm$ 0.01 \\
& GRU & 7.40 $\pm$ 0.31 & 5.61 $\pm$ 0.28 & 0.34 $\pm$ 0.06 \\
& LightGBM & 5.98 $\pm$ 0.00 & 4.39 $\pm$ 0.00 & 0.57 $\pm$ 0.00 \\
& Transformer & 7.02 $\pm$ 0.48 & 5.14 $\pm$ 0.31 & 0.41 $\pm$ 0.08 \\
\midrule
\multirow{3}{*}{Multistage Methods}
& TimeXer & 7.98 $\pm$ 0.14 & 6.28 $\pm$ 0.10 & 0.24 $\pm$ 0.03 \\
& PEGNN-SS & 9.74 $\pm$ 0.53 & 7.51 $\pm$ 0.70 & $-0.10 \pm 0.12$ \\
& MultistageNet & 5.75 $\pm$ 0.24 & 4.19 $\pm$ 0.08 & 0.61 $\pm$ 0.03 \\
\midrule
Ours & DAMF & \best{5.24 $\pm$ 0.13} & \best{3.60 $\pm$ 0.13} & \best{0.67 $\pm$ 0.02} \\
\bottomrule
\end{tabular}
}
\tabnote{\raggedright
Results are reported as mean $\pm$ standard deviation over three random seeds.
RMSE and MAE are lower-is-better metrics, while $R^2$ is a higher-is-better metric.\par}
\label{tab:liquidsugar_prediction}
\end{table}

Table~\ref{tab:liquidsugar_prediction} reports the final quality prediction results on the LiquidSugar dataset. Among traditional machine-learning baselines, LightGBM achieves the best overall performance, followed by Random Forest and XGBoost, suggesting that tree-based ensemble methods can capture part of the nonlinear relationship between process variables and final quality. By contrast, generic deep sequential models, including LSTM, GRU, and Transformer, do not consistently outperform traditional baselines, which suggests that directly applying sequence models to timestamp-aligned multi-stage observations is insufficient for this manufacturing prediction task. Multi-stage baselines show mixed performance. TimeXer and PEGNN-SS do not outperform the strongest traditional baselines, whereas MultistageNet achieves the best baseline performance across all three metrics. This comparison indicates that explicitly modeling the stage structure is useful, but representing the production flow alone is still insufficient. Without explicit modeling of dynamic inter-stage delays, upstream and downstream process states may remain imperfectly matched. DAMF achieves the best performance across all three metrics. Compared with the strongest baseline, MultistageNet, DAMF consistently reduces prediction errors and improves the explained variance of the final quality response. These improvements show that estimating inter-stage delays and using them to guide multi-stage fusion provides additional predictive benefit beyond modeling the stage structure alone.

\subsection{Lag recovery on controlled-delay benchmarks}

\subsubsection{Quantitative comparison}
\begin{figure}[!t]
\centering
\includegraphics[width=0.92\textwidth]{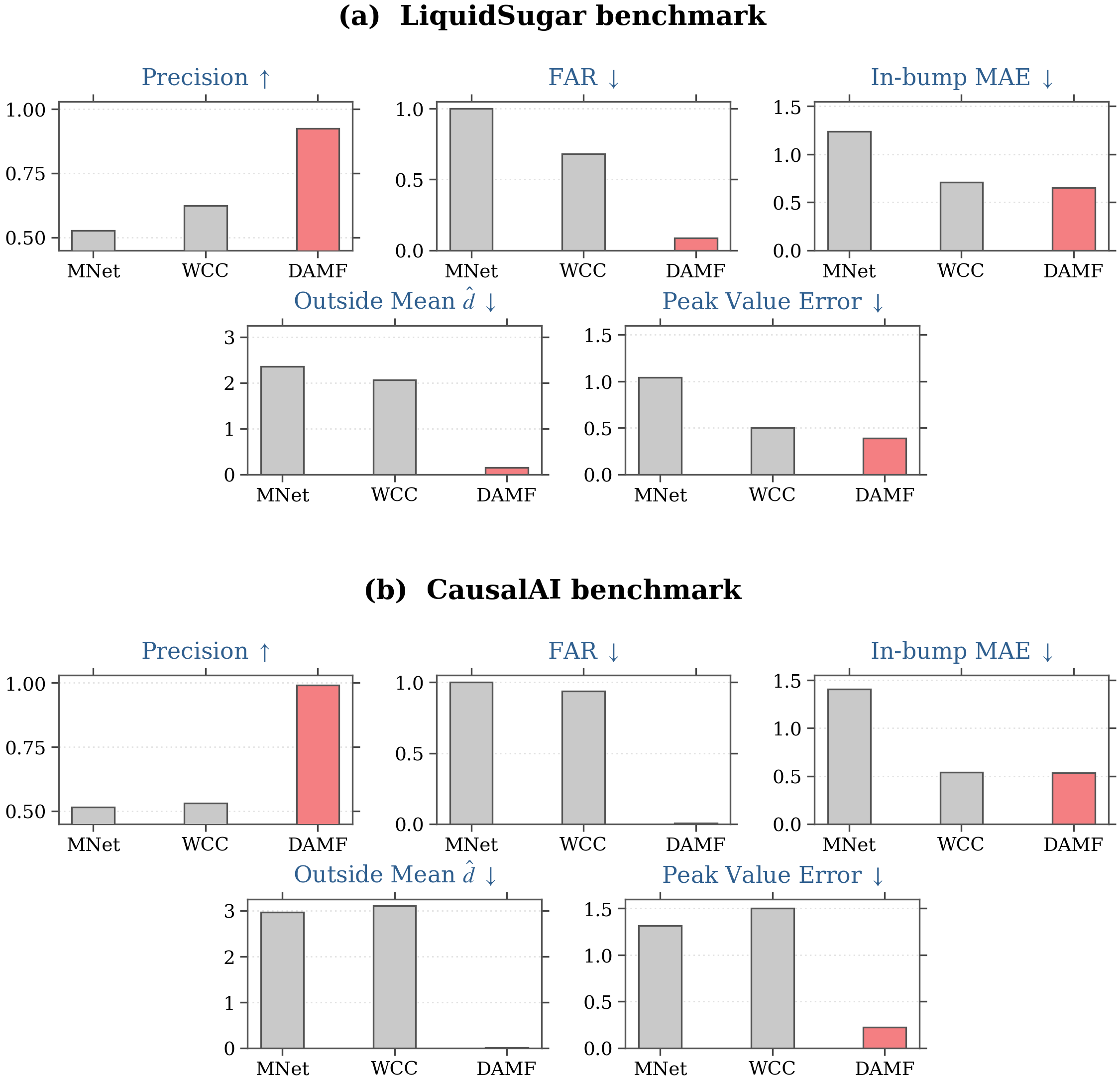}
\caption{Lag-recovery comparison on the two controlled-delay benchmarks. Each panel reports one lag-recovery metric for MNet, WCC, and DAMF. MNet uses the hidden-lag output of MultistageNet, WCC combines window-based cross-correlation with a conventional Viterbi decoder, and DAMF denotes the proposed lag-recovery pipeline and is highlighted in red.}
\label{fig:lag_recovery_barplots}
\end{figure}

We compare DAMF with MNet and WCC on the two controlled-delay benchmarks
using the metrics defined in Section~\ref{sec:evaluation_metrics}:
precision, FAR, In-bump MAE, Outside mean delay, and Peak Value Error. Figure~\ref{fig:lag_recovery_barplots} summarizes the
quantitative results on the two controlled-delay benchmarks.
DAMF achieves higher precision and lower FAR than MNet
and WCC on both benchmarks.
On LiquidSugar, it also reduces in-bump MAE, outside
mean delay, and peak value error.
On CausalAI, DAMF achieves near-perfect precision and
a low FAR. Its in-bump MAE is close to that of WCC,
whereas its outside mean delay and peak value error
are substantially lower.
These results indicate that DAMF improves both the
detection of delayed intervals and the estimation of
delay magnitudes under controlled-delay settings.

\subsubsection{Representative lag recovery examples}
\begin{figure}[!htbp]
\centering
\includegraphics[width=0.92\textwidth]{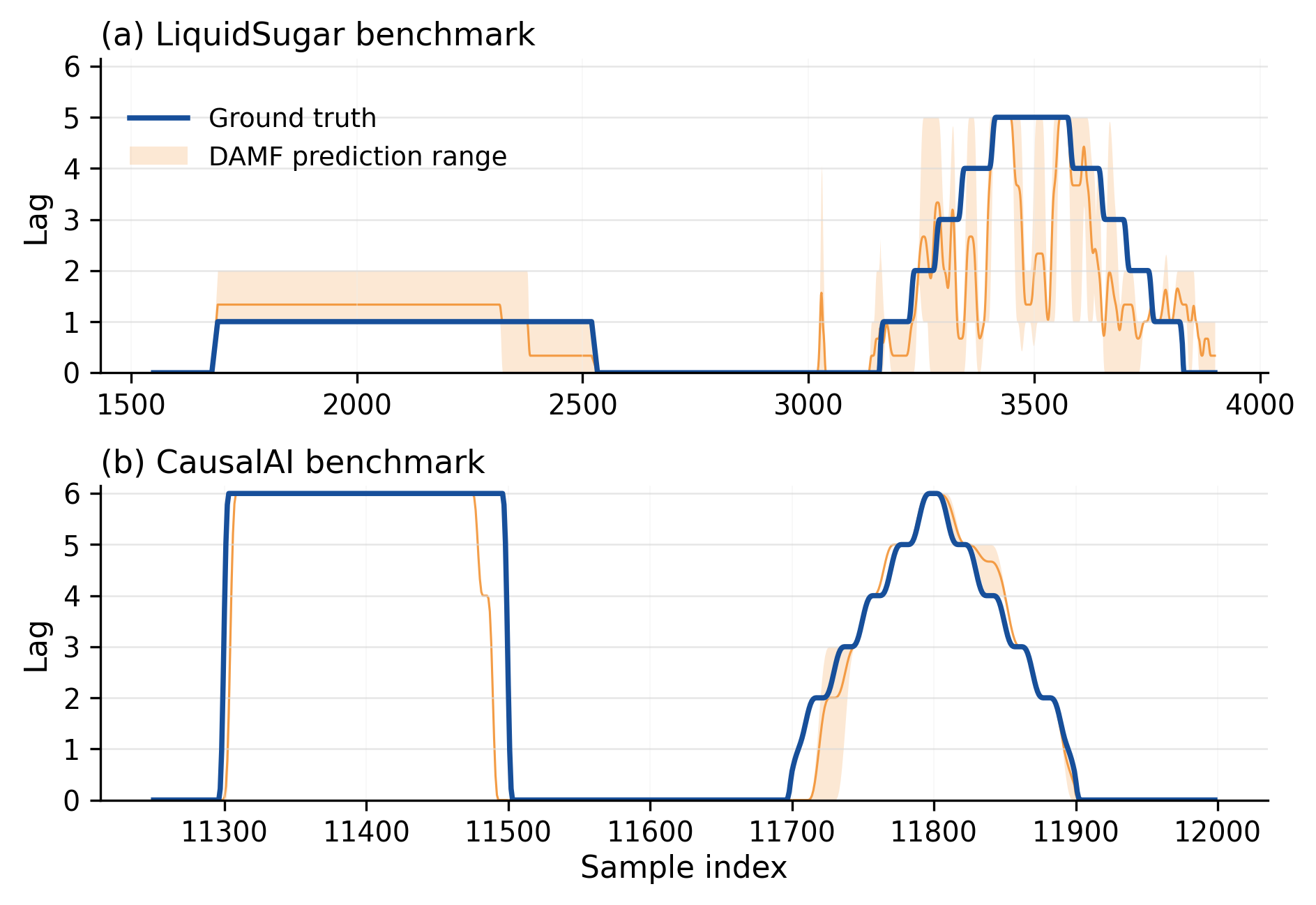}
\caption{Representative lag-recovery examples on the controlled-delay benchmarks. The blue curves denote the ground-truth lag paths, while the orange curves and shaded regions represent the mean decoded paths and their ranges across random seeds, respectively. Each benchmark contains both block-shaped and bump-shaped delay patterns.}
\label{fig:lag_recovery_controlled}
\end{figure}

Figure~\ref{fig:lag_recovery_controlled} provides qualitative examples of the decoded lag paths produced by DAMF on the two controlled-delay benchmarks. The blue curve denotes the ground-truth lag path. Let $\hat d_t^{(r)}$ denote the decoded delay at sample index $t$ in the $r$-th random-seed run, with $r=1,\ldots,R$. The predictions from all runs are aligned by sample index. At each index $t$, the orange curve shows the mean $\bar d_t=R^{-1}\sum_{r=1}^{R}\hat d_t^{(r)}$, while the lower and upper boundaries of the shaded region are $\min_{1\leq r\leq R}\hat d_t^{(r)}$ and $\max_{1\leq r\leq R}\hat d_t^{(r)}$, respectively. 

These summaries are computed from the decoded lag paths
of the individual runs. The shaded band therefore shows the observed variation across random seeds, rather than a confidence interval;
a wider band indicates greater disagreement among runs. For LiquidSugar, the recovered paths capture the main injected delay intervals and follow the overall ground-truth trajectory. Both block-shaped and bump-shaped delay structures are retained, with discrepancies occurring mainly near transition boundaries and rapidly changing lag values. For CausalAI, the decoded paths show closer agreement with the ground truth. The sustained block and the smooth bump are both recovered, including their onset, peak, and decay regions. This pattern is consistent with the cleaner source--target dependence in the CausalAI benchmark. Additional region-level lag-recovery examples are provided in Appendix~\ref{app:additional_lag_recovery}.

\subsection{Ablation studies}
The following ablations examine the lag-predictor architecture,
training objective, temporal decoding, and lag prior.
All lag-recovery metrics follow the definitions in
Section~\ref{sec:evaluation_metrics}.

\subsubsection{Structural ablation of the lag predictor}
The structural ablation evaluates three key design choices in the lag predictor: source-target lag matching, lag-specific source value extraction, and the selector auxiliary head. These designs are introduced to enable explicit delay comparison, preserve source information under each candidate lag, and provide auxiliary supervision for candidate ranking, respectively. Table~\ref{tab:stda_structure_ablation} shows that weakening any of these designs leads to worse lag recovery performance. When source-target lag matching is removed, Occ F1 drops substantially and Lag MAE increases, indicating that directly comparing upstream and downstream temporal windows is important for localizing delays. Removing lag-specific source value extraction causes the largest degradation, especially in large-lag recall and collapse-to-zero errors. This suggests that retaining source information at each candidate lag is critical for distinguishing nonzero delays. Removing the selector auxiliary head also reduces performance, although the drop is relatively smaller, which indicates that it mainly serves as an auxiliary regularizer for candidate ranking rather than the primary source of lag recovery.
\begin{table}[!h]
\small
\setlength{\tabcolsep}{3.5pt}
\renewcommand{\arraystretch}{1.12}
\tbl{Structural ablation of the lag predictor. The best result for each metric is shown in bold.}
{\begin{tabular}{lccccc}
\toprule
Setting & Occ F1 $\uparrow$ & Lag MAE $\downarrow$ & $\pm 1$ Acc $\uparrow$ & Large-lag Recall $\uparrow$ & Collapse to 0 $\downarrow$ \\
\midrule
Full lag predictor
& \textbf{0.91 $\pm$ 0.02}
& \textbf{0.65 $\pm$ 0.13}
& \textbf{0.90 $\pm$ 0.02}
& \textbf{0.76 $\pm$ 0.05}
& \textbf{0.03 $\pm$ 0.03} \\

w/o Match
& 0.60 $\pm$ 0.06
& 2.34 $\pm$ 0.19
& 0.64 $\pm$ 0.04
& 0.48 $\pm$ 0.11
& 0.45 $\pm$ 0.08 \\

w/o LagVal
& 0.42 $\pm$ 0.05
& 2.45 $\pm$ 0.08
& 0.64 $\pm$ 0.02
& 0.33 $\pm$ 0.00
& 0.64 $\pm$ 0.02 \\

w/o Selector
& 0.86 $\pm$ 0.10
& 0.90 $\pm$ 0.32
& 0.87 $\pm$ 0.04
& 0.67 $\pm$ 0.11
& 0.14 $\pm$ 0.16 \\
\bottomrule
\end{tabular}}
\tabnote{w/o Match removes source-target lag matching; w/o LagVal removes the lag-specific source value; w/o Selector removes the selector auxiliary head.}
\label{tab:stda_structure_ablation}
\end{table}

\subsubsection{Loss ablation of the lag predictor}
\label{subsec:stda_loss_ablation}
We further investigate the contribution of each loss term in the lag predictor training objective. 
The ablation settings are constructed by removing one loss term at a time from the full objective, including the relaxed-TopN loss, the positive-lag CE loss, the large-occurrence recall loss, and the selector loss.  In addition, we include a simplified variant that retains only the occurrence loss and the positive-lag CE loss, so as to examine whether standard classification supervision is sufficient for reliable lag recovery. Table~\ref{tab:stda_loss_ablation} shows that the full lag-predictor objective achieves the strongest overall performance. 
It obtains the best results in Occ F1, Lag MAE, $\pm 1$ Acc, and Collapse to 0, while maintaining competitive large-lag recall. 
Although the simplified occurrence-plus-CE setting still yields reasonable results, it noticeably reduces large-lag recall and increases the collapse-to-zero rate. This indicates that basic classification supervision alone is insufficient to protect difficult large-lag regions, especially when the model needs to distinguish true nonzero delays from no-lag cases.

Among the individual loss terms, the relaxed-TopN loss mainly improves candidate coverage and helps preserve plausible large-lag candidates. 
When this term is removed, the recovered lag path becomes less stable and the overall lag estimation accuracy deteriorates. 
The positive-lag CE loss is essential for estimating the lag magnitude; removing it leads to the largest increase in Lag MAE and a substantial drop in $\pm 1$ Acc. 
The large-occurrence recall loss further prevents large delays from being misclassified as no lag or small lag, as reflected by the degradation in large-lag recall and collapse-related metrics. 
The selector loss has a relatively moderate but consistent effect, suggesting that it serves primarily as an auxiliary regularizer for candidate ranking rather than as the dominant source of lag recovery.
\label{subsec:stda_loss_ablation}
\begin{table*}[!h]
\centering
\setlength{\tabcolsep}{3.2pt}
\renewcommand{\arraystretch}{1.10}
\footnotesize

\resizebox{\textwidth}{!}{%
\begin{tabular}{lccccc}
\toprule
Setting
& Occ F1 $\uparrow$
& Lag MAE $\downarrow$
& $\pm 1$ Acc $\uparrow$
& Large-lag Recall $\uparrow$
& Collapse to 0 $\downarrow$ \\
\midrule

Full lag-prediction objective
& \textbf{0.91 $\pm$ 0.02}
& \textbf{0.65 $\pm$ 0.13}
& \textbf{0.90 $\pm$ 0.02}
& 0.76 $\pm$ 0.05
& \textbf{0.03 $\pm$ 0.03} \\

Occurrence + positive CE only
& 0.90 $\pm$ 0.02
& 0.82 $\pm$ 0.15
& 0.87 $\pm$ 0.02
& 0.68 $\pm$ 0.05
& 0.20 $\pm$ 0.09 \\

w/o relaxed-TopN loss
& 0.89 $\pm$ 0.04
& 0.94 $\pm$ 0.15
& 0.86 $\pm$ 0.02
& \textbf{0.76 $\pm$ 0.13}
& 0.10 $\pm$ 0.08 \\

w/o positive-lag CE loss
& 0.64 $\pm$ 0.11
& 2.77 $\pm$ 0.08
& 0.60 $\pm$ 0.01
& 0.14 $\pm$ 0.02
& 0.65 $\pm$ 0.04 \\

w/o large-occurrence recall loss
& 0.83 $\pm$ 0.07
& 1.32 $\pm$ 0.23
& 0.82 $\pm$ 0.03
& 0.62 $\pm$ 0.16
& 0.23 $\pm$ 0.17 \\

w/o selector loss
& 0.86 $\pm$ 0.05
& 0.91 $\pm$ 0.39
& 0.85 $\pm$ 0.05
& 0.65 $\pm$ 0.15
& 0.12 $\pm$ 0.09 \\

\bottomrule
\end{tabular}}

\vspace{3pt}

\refstepcounter{table}
\label{tab:stda_loss_ablation}
\noindent
\begin{minipage}{\textwidth}
\raggedright
\textbf{Table~\thetable.} Loss ablation for lag recovery. The first row uses the complete lag-prediction objective. Each row removes one loss component from the full objective, except ``Occurrence + positive-lag CE only''. The best numerical result for each metric is shown in bold.
\end{minipage}

\end{table*}

\subsubsection{Effect of Viterbi decoding}

After the lag predictor produces raw edge-level lag emissions, the final lag path can be obtained either by pointwise raw-argmax decisions or by structured temporal decoding. 
The raw-argmax strategy treats each sample independently and therefore does not explicitly account for the temporal persistence of manufacturing delays. 
In contrast, Viterbi decoding imposes structured continuity on the recovered lag sequence, which is consistent with the fact that process delays usually persist over continuous intervals rather than switching arbitrarily at each time point.

\begin{figure}[!t]
\centering
\includegraphics[width=0.95\textwidth]{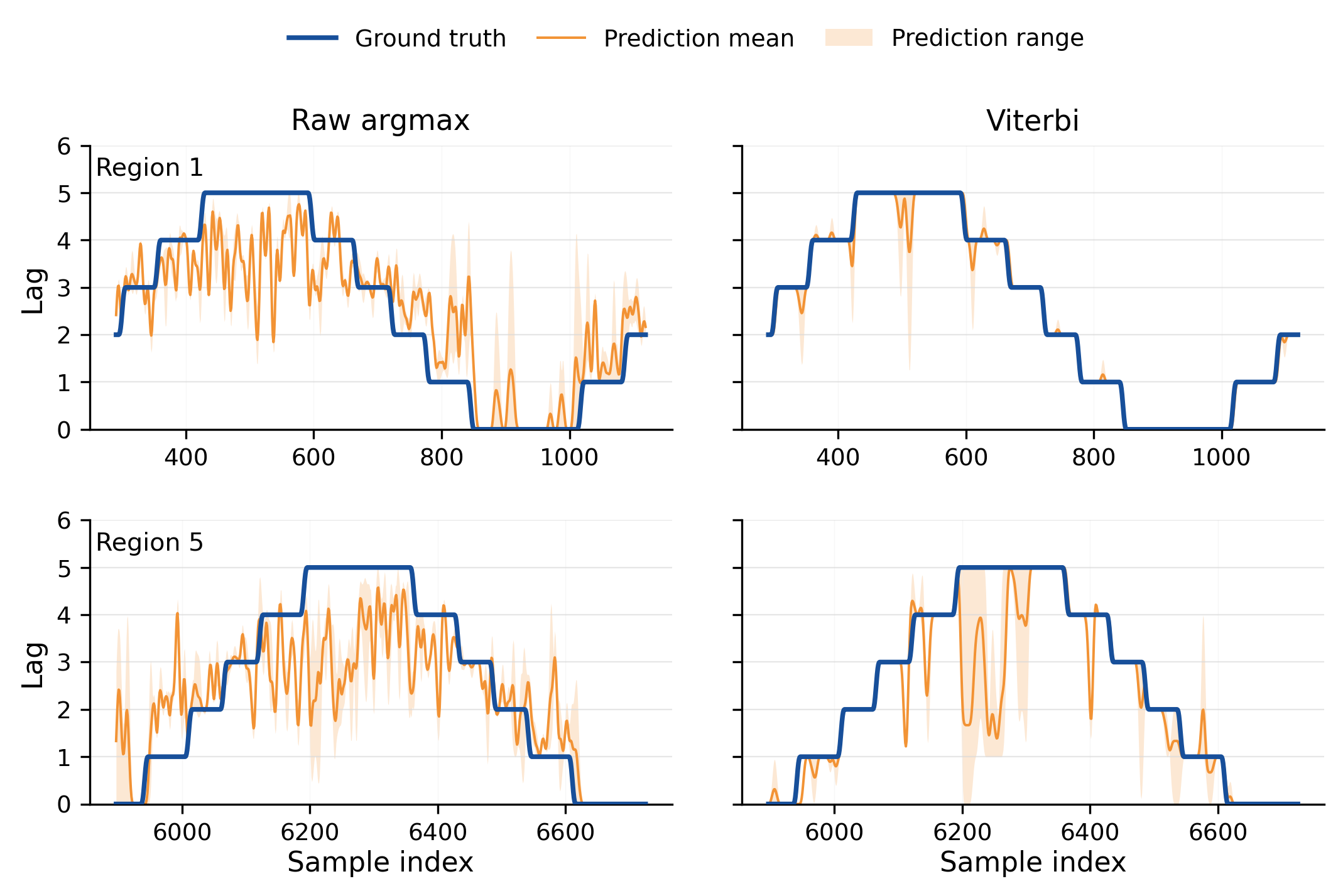}
\caption{Path-level comparison between raw pointwise argmax predictions and Viterbi decoding on representative test regions. The left
column shows raw argmax predictions before decoding, whereas the right column
shows the lag paths after Viterbi decoding. The blue curve
denotes the ground-truth lag path, the orange curve denotes the prediction
mean across random seeds, and the shaded region denotes the corresponding
prediction range.}
\label{fig:viterbi_before_after}
\end{figure}

Table~\ref{tab:viterbi_ablation} compares raw-argmax decoding with Viterbi decoding. 
The decoder substantially improves the main accuracy-related metrics, including Acc, $\pm 1$ Acc, Large-lag Recall, and Lag MAE. 
It also reduces Collapse to small, indicating that structured decoding helps suppress unstable small-lag predictions. 
Raw argmax achieves a slightly lower Collapse to 0 value; however, the absolute values of this metric are small in both settings. 
Therefore, the overall comparison demonstrates that Viterbi decoding improves lag recovery accuracy and produces more stable lag paths.  Figure~\ref{fig:viterbi_before_after} further provides a path-level illustration of this effect. Raw-argmax predictions are able to capture the broad delayed regions, but they contain local spikes, short-lived drops, and noticeable seed-to-seed variability. 
After decoding, the recovered paths become less fragmented and better aligned with the duration of the underlying delay regions. 
This qualitative observation is consistent with the aggregate improvements reported in Table~\ref{tab:viterbi_ablation}.
\begin{table*}[t]
\centering
\setlength{\tabcolsep}{3.8pt}
\renewcommand{\arraystretch}{1.10}
\footnotesize
\caption{Effect of Viterbi decoding on lag recovery. The best numerical result for each metric is shown in bold.}
\label{tab:viterbi_ablation}
\resizebox{\textwidth}{!}{%
\begin{tabular}{lcccccc}
\toprule
Setting
& Acc $\uparrow$
& $\pm 1$ Acc $\uparrow$
& Lag MAE $\downarrow$
& Large-lag Recall $\uparrow$
& Collapse to 0 $\downarrow$
& Collapse to small $\downarrow$ \\
\midrule

Raw argmax
& 0.65 $\pm$ 0.04
& 0.79 $\pm$ 0.04
& 1.03 $\pm$ 0.12
& 0.55 $\pm$ 0.04
& \textbf{0.02 $\pm$ 0.02}
& 0.44 $\pm$ 0.03 \\

Viterbi
& \textbf{0.76 $\pm$ 0.05}
& \textbf{0.90 $\pm$ 0.02}
& \textbf{0.65 $\pm$ 0.13}
& \textbf{0.76 $\pm$ 0.05}
& 0.03 $\pm$ 0.03
& \textbf{0.22 $\pm$ 0.03} \\

\bottomrule
\end{tabular}}
\end{table*}
Taken together, these experiments provide consistent evidence for the proposed delay-integrated design. 
DAMF improves final quality prediction, lag predictor recovers interpretable time-varying lag patterns on controlled-delay benchmarks, and the ablation studies verify the necessity of the lag prior, the training objective, and the duration-aware decoder. 
The performance gain therefore arises from the joint effect of lag estimation and delay-aware fusion, rather than from any single part alone.

\subsubsection{Effect of the lag prior on final quality prediction}

This ablation examines the contribution of the estimated lag prior to final quality prediction. We compare DAMF without the lag prior against variants that incorporate the pretrained lag predictor's soft distribution into DelayAlignment. All variants use the same forecasting architecture and training protocol, with only the prior weight varying. We evaluated different prior weights on the train set and adopted $\lambda=0.4$ in the main experiments, as it achieved the best overall performance across RMSE, MAE, and $R^2$.

\begin{figure}[!t]
\centering
\includegraphics[width=0.92\textwidth]{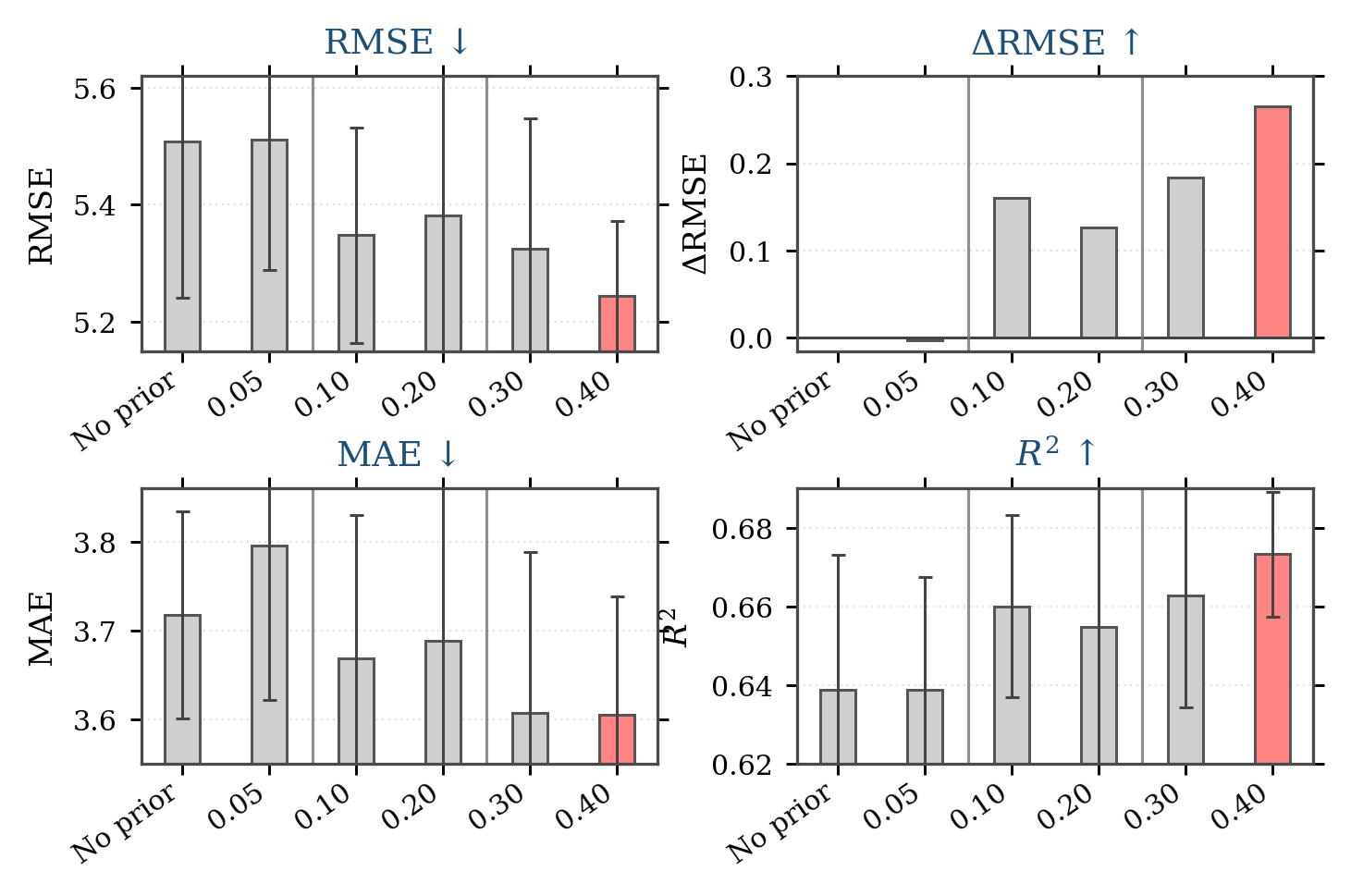}
\caption{Sensitivity analysis of the lag prior weight. The four panels report RMSE, $\Delta$RMSE, MAE, and $R^2$ under different prior-weight settings.}
\label{fig:app-prior-weight-sensitivity}
\end{figure}

Figure~\ref{fig:app-prior-weight-sensitivity} presents the results under different prior weights. Compared with the no-prior baseline, most lag-prior settings yield lower RMSE and MAE and higher $R^2$. The improvement is marginal at the smallest weight, while $\lambda=0.4$ gives the best results across all three metrics. These results suggest that the estimated edge-level lag distributions provide useful information for cross-stage alignment and can improve final quality prediction.
Additional region-level lag recovery examples are provided in Appendix~\ref{app:additional_lag_recovery}.

\section{Conclusion}

This paper presented DAMF, a delay-aware framework for multi-stage manufacturing quality prediction. The central idea is to treat inter-stage propagation delay as prediction-relevant information, rather than relying on fixed timestamp alignment or leaving lag estimation as a separate diagnostic analysis. By integrating delay estimation into the forecasting process, DAMF improves the connection between upstream process histories and downstream quality outcomes while also providing interpretable temporal propagation patterns across production stages.

The experimental results show that delay information plays a practical role in multi-stage quality prediction. DAMF outperforms the compared baselines on final quality prediction, and the lag predictor recovers more accurate and stable lag paths on controlled-delay benchmarks. These findings indicate that the estimated inter-stage delays can serve not only as interpretable diagnostic outputs, but also as useful temporal guidance for aligning upstream and downstream process representations.

Ablation studies further clarify the source of the performance gains. Removing the lag prior weakens final quality prediction, simplifying the lag-predictor architecture or its training objective degrades lag recovery, and replacing structured decoding with pointwise decisions leads to less stable lag paths. These results support two distinct uses of the estimated lag distributions: guiding cross-stage alignment for quality prediction and recovering temporally coherent lag paths
for interpretation.

Several directions remain for future work. First, the current framework focuses on adjacent-stage delay modeling, whereas more complex manufacturing systems may involve non-adjacent interactions, branching process structures, or feedback loops. Extending DAMF to more general process graphs would broaden its applicability. Second, the current delay representation is based on a discrete candidate lag space. Continuous or distributional delay modeling may provide a more flexible way to describe propagation patterns.

\bibliographystyle{tfcad}
\bibliography{interacttfpsample}
\appendix

\makeatletter
\@addtoreset{figure}{section}
\makeatother
\renewcommand{\thefigure}{\Alph{section}\arabic{figure}}

\section{Additional lag recovery examples}
\label{app:additional_lag_recovery}

\begin{figure}[!htbp]
\centering
\subfigure[LiquidSugar dataset with artificially injected lag.]{
\includegraphics[width=\textwidth]{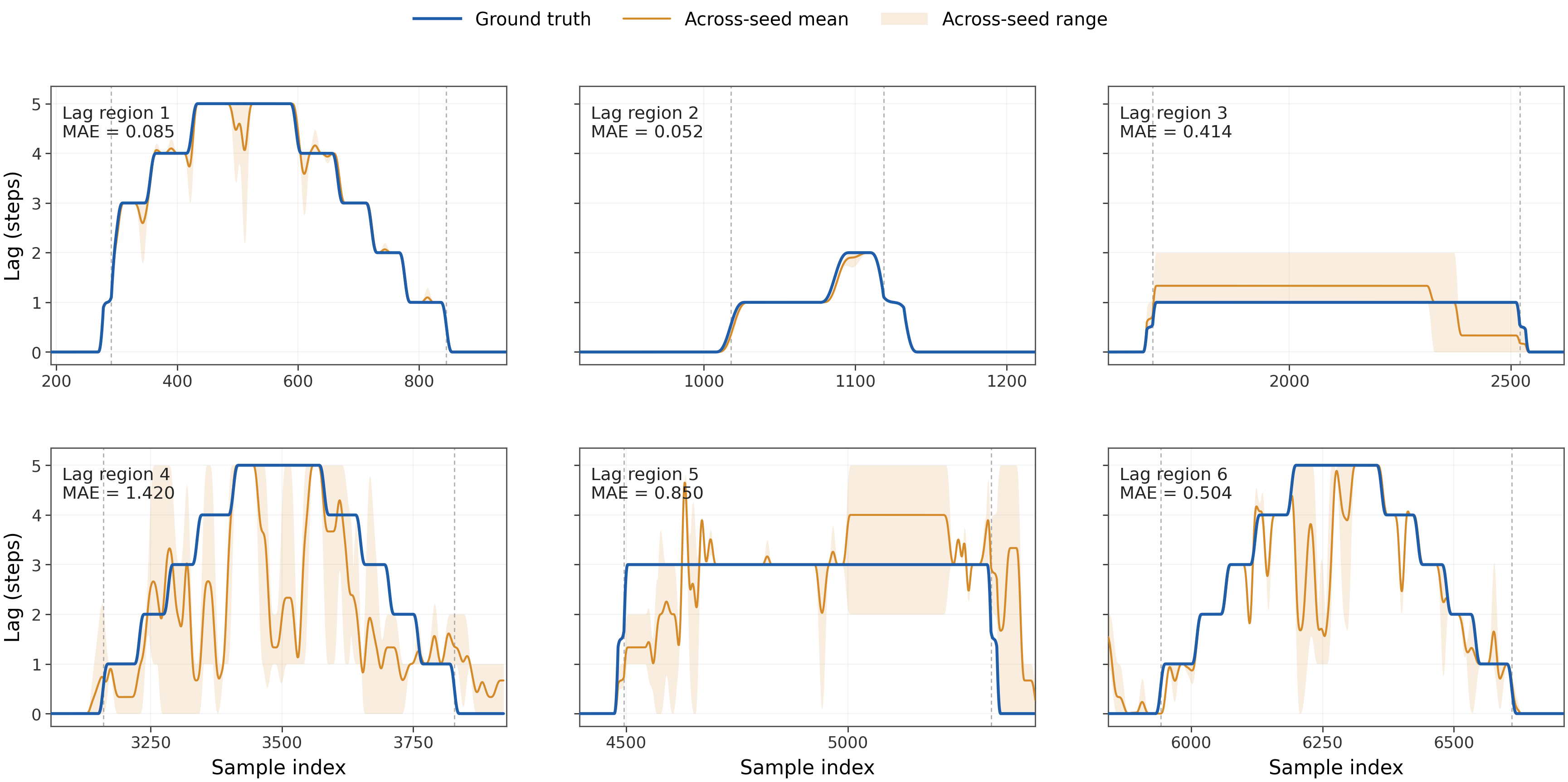}}
\\[4pt]
\subfigure[CausalAI dataset.]{
\includegraphics[width=\textwidth]{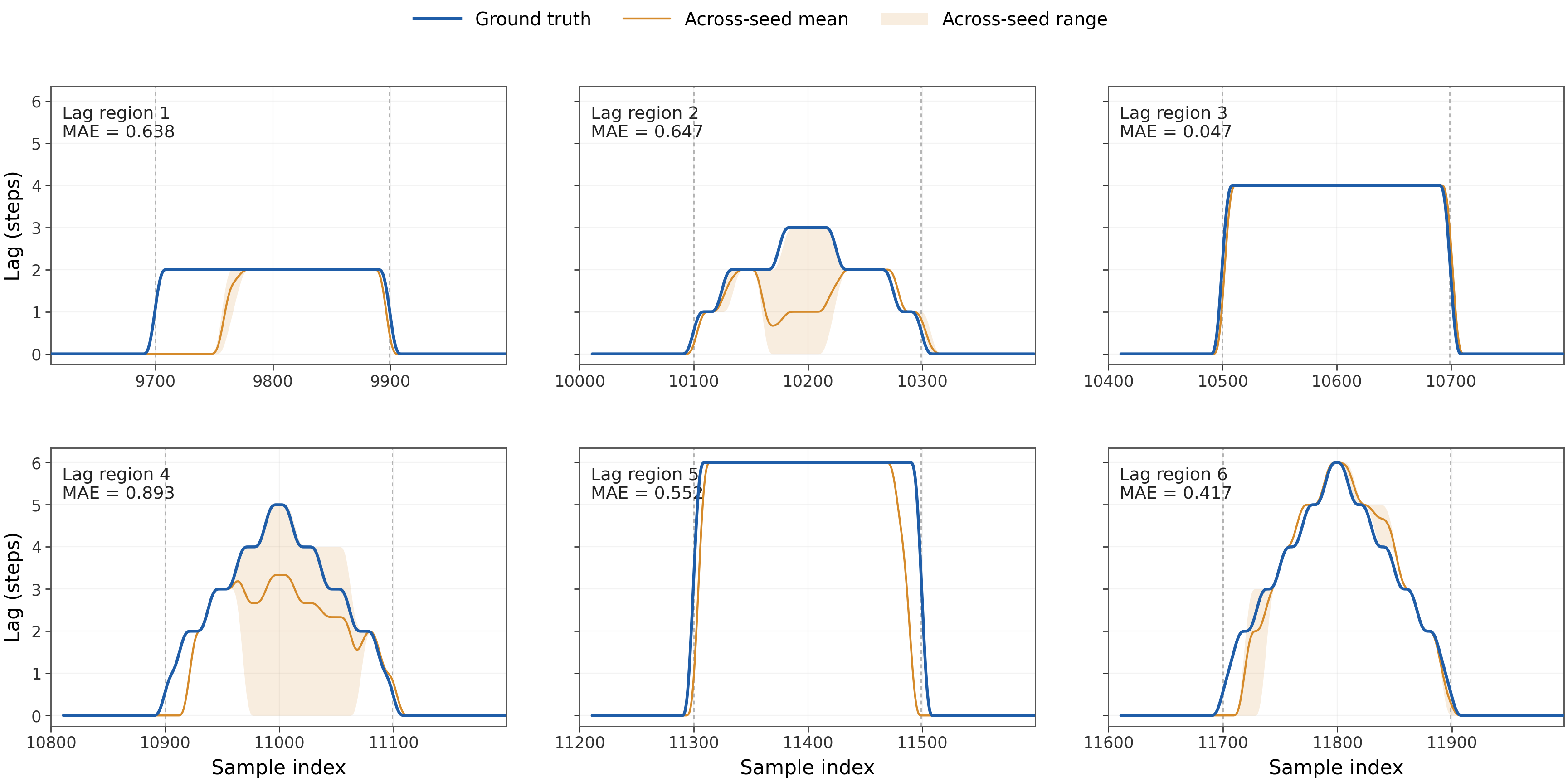}}
\caption{Lag recovery results of the edge-level lag predictor on two controlled-delay datasets. The blue line denotes the ground-truth lag path, the orange line denotes the mean Viterbi-decoded lag path across seeds, and the shaded region denotes the corresponding range across
seeds 42, 43, and 44.}
\label{fig:lag_recovery_controlled_all}
\end{figure}

\section{Additional lag path recovery results}
\label{app:lag-path-recovery}
\begin{figure}[!t]
    \centering
    \includegraphics[width=\textwidth]{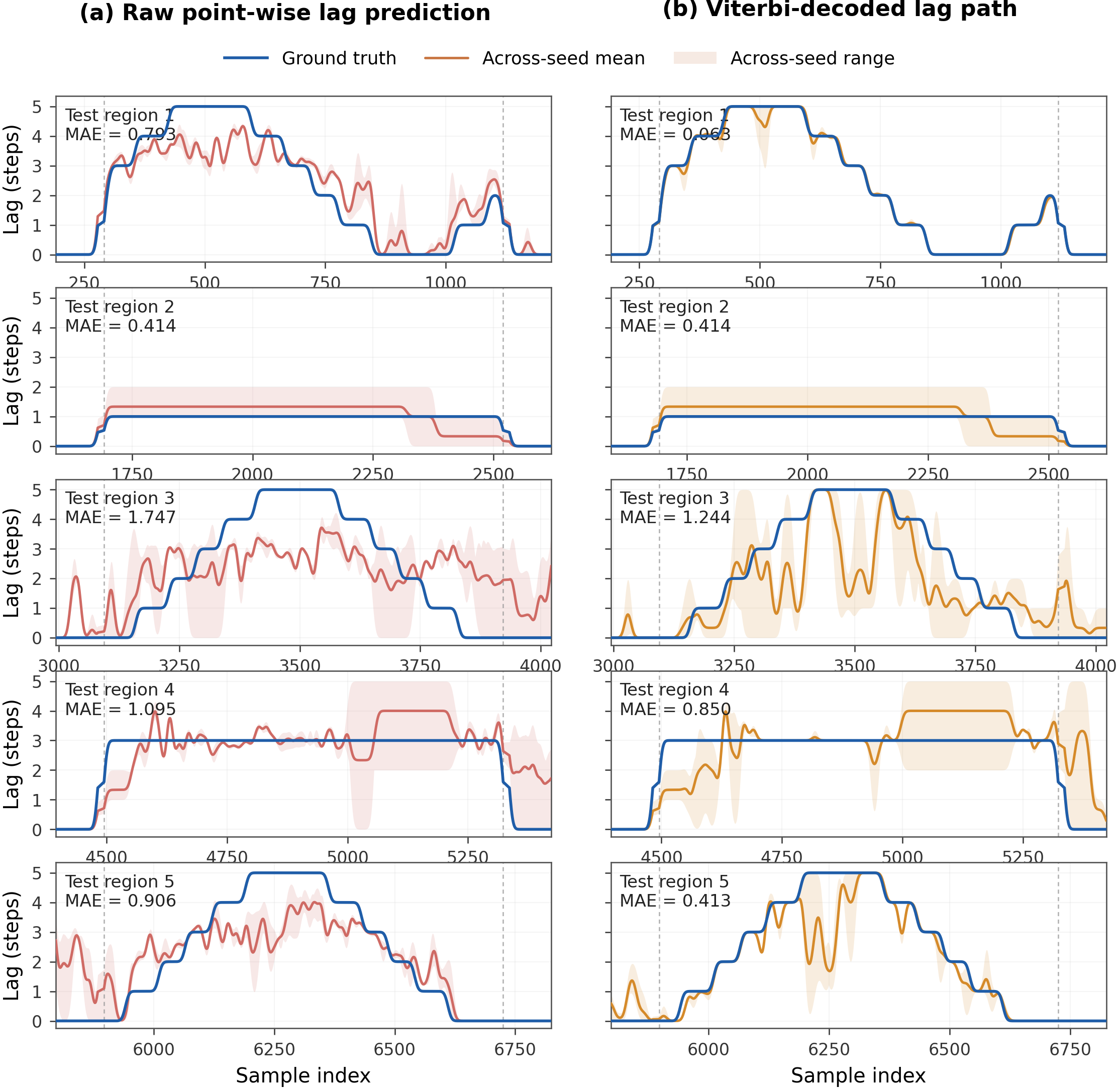}
    \caption{Effect of Viterbi decoding on lag-path recovery across five test regions. The left and right columns show the raw point-wise lag predictions and the corresponding Viterbi-decoded lag paths, respectively. Solid colored curves and shaded bands denote the mean and range across three random seeds, while blue curves represent the ground-truth lag paths. Vertical dashed lines delimit the evaluation intervals, and the reported MAE is computed within each interval.}
    \label{fig:viterbi-comparison}
\end{figure}

\section{Detailed lag-predictor objective}
\label{app:lag-losses}

For clarity, we present the objective for a fixed adjacent-stage edge and
omit the edge superscript \(g\). Let \(\mathcal{B}\) denote a nonempty
mini-batch, let
\(\ell_t\in\mathcal{K}=\{0,1,\ldots,K\}\) denote the ground-truth lag of
sample \(t\), and define
\[
r_t=\mathbb{I}(\ell_t>0).
\]
The occurrence head outputs \(o_t=\Pr(z_t>0)\), and the positive-lag head
outputs
\[
\pi_{\mathrm{pos},t}(k)
=
\Pr(z_t=k\mid z_t>0),
\qquad
k\in\mathcal{K}_{+},
\qquad
\mathcal{K}_{+}=\{1,\ldots,K\}.
\]
For any index set \(\mathcal{A}\), define
\[
D(\mathcal{A})=\max\{1,|\mathcal{A}|\}.
\]
Thus, a loss whose contributing set is empty evaluates to zero. The occurrence head is trained using a focal-style binary classification
loss:
\begin{equation}
\mathcal{L}_{\mathrm{occ}}
=
-\frac{1}{|\mathcal{B}|}
\sum_{t\in\mathcal{B}}
\left[
\alpha r_t(1-o_t)^{\gamma}
\log(o_t+\epsilon)
+
(1-\alpha)(1-r_t)o_t^{\gamma}
\log(1-o_t+\epsilon)
\right],
\label{eq:app-lag-occ}
\end{equation}
where \(\alpha\in[0,1]\) controls class weighting,
\(\gamma\geq0\) is the focal parameter, and \(\epsilon>0\) ensures
numerical stability. Let
\[
\mathcal{P}
=
\{t\in\mathcal{B}:\ell_t>0\}
\]
denote the set of positive-lag samples. The positive-lag head is trained
using the conditional cross-entropy loss
\begin{equation}
\mathcal{L}_{\mathrm{pos}}
=
-\frac{1}{D(\mathcal{P})}
\sum_{t\in\mathcal{P}}
\log\!\left(
\pi_{\mathrm{pos},t}(\ell_t)+\epsilon
\right).
\label{eq:app-lag-pos}
\end{equation}
For each \(t\in\mathcal{P}\), define the set of incorrect positive-lag
candidates as
\[
\mathcal{N}_t
=
\mathcal{K}_{+}\setminus\{\ell_t\}.
\]
Let
\[
s_{t,[1]}^{(-)}
\geq
s_{t,[2]}^{(-)}
\geq
\cdots
\geq
s_{t,[K-1]}^{(-)}
\]
be the matching scores
\(\{s_t(k):k\in\mathcal{N}_t\}\) sorted in descending order, and let
\[
s_{t,(N)}^{(-)}=s_{t,[N]}^{(-)},
\qquad
1\leq N\leq K-1.
\]
The relaxed Top-\(N\) loss is
\begin{equation}
\mathcal{L}_{\mathrm{TopN}}
=
\frac{1}{D(\mathcal{P})}
\sum_{t\in\mathcal{P}}
\left[
m+s_{t,(N)}^{(-)}-s_t(\ell_t)
\right]_{+},
\qquad
[x]_{+}=\max\{0,x\},
\label{eq:app-lag-topn}
\end{equation}
where \(m>0\) is a margin. This term encourages the ground-truth lag to
remain among the leading candidates without requiring it to have the
largest matching score. To discourage large positive lags from being classified as zero, define
\[
\mathcal{P}_{\mathrm{large}}
=
\left\{
t\in\mathcal{B}:
\ell_t\geq K_{\mathrm{large}}
\right\},
\qquad
K_{\mathrm{large}}\in\{1,\ldots,K\}.
\]
The corresponding occurrence loss is
\begin{equation}
\mathcal{L}_{\mathrm{large}}
=
-\frac{1}{D(\mathcal{P}_{\mathrm{large}})}
\sum_{t\in\mathcal{P}_{\mathrm{large}}}
\log(o_t+\epsilon).
\label{eq:app-lag-large}
\end{equation}
Let
\[
\mathcal{C}_t
=
(k_{t,1},\ldots,k_{t,N})
\]
denote the \(N\) highest-scoring candidates in
\(\mathcal{K}_{+}\), ordered such that
\[
s_t(k_{t,1})
\geq
s_t(k_{t,2})
\geq
\cdots
\geq
s_t(k_{t,N}).
\]
Let \(a_t(j)\) be the selector probability assigned to candidate
position \(j\), satisfying
\[
a_t(j)\geq0,
\qquad
\sum_{j=1}^{N}a_t(j)=1.
\]
For samples whose ground-truth lag is retained, define
\[
\mathcal{S}
=
\left\{
t\in\mathcal{P}:
\ell_t\in\mathcal{C}_t
\right\},
\qquad
j_t^{*}
=
\operatorname{index}_{\mathcal{C}_t}(\ell_t).
\]
The selector loss is
\begin{equation}
\mathcal{L}_{\mathrm{sel}}
=
-\frac{1}{D(\mathcal{S})}
\sum_{t\in\mathcal{S}}
\log\!\left(
a_t(j_t^{*})+\epsilon
\right).
\label{eq:app-lag-sel}
\end{equation}
The selector provides auxiliary ranking supervision only and is not used
to construct the lag distribution at inference. The complete lag-predictor objective is
\begin{equation}
\mathcal{L}_{\mathrm{lag}}
=
\mathcal{L}_{\mathrm{occ}}
+
\mathcal{L}_{\mathrm{pos}}
+
\mathcal{L}_{\mathrm{TopN}}
+
\lambda_{\mathrm{large}}
\mathcal{L}_{\mathrm{large}}
+
\lambda_{\mathrm{sel}}
\mathcal{L}_{\mathrm{sel}},
\label{eq:app-lag-total}
\end{equation}
where
\(\lambda_{\mathrm{large}},\lambda_{\mathrm{sel}}\geq0\) control the
auxiliary terms. The occurrence, positive-lag, and relaxed Top-\(N\)
terms have unit coefficients in this formulation.

\end{document}